\documentclass[10pt,journal,compsoc]{IEEEtran}
\usepackage{times}
\usepackage{epsfig}
\usepackage{graphicx,subcaption}
\usepackage{color}
\usepackage{dirtytalk}
\usepackage{comment}
\usepackage{wrapfig}
\usepackage{caption}
\usepackage{amsmath}
\usepackage{amssymb}
\usepackage{lipsum,booktabs}
\usepackage[colorlinks]{hyperref}

\newcommand{\todo}[1]{}
\renewcommand{\todo}[1]{{\color{red} TODO: {#1}}}
\newcommand{\LMR}[1]{{$MR^{-2}$}}

\newcommand{\caltech}[1]{{Caltech~\cite{dollar2012pedestrian}}}
\newcommand{\cityperson}[1]{{CityPersons~\cite{zhang2017citypersons}}}
\newcommand{\ecp}[1]{{ECP~\cite{braun2018eurocity}}}
\newcommand{\widerperson}[1]{{Wider Pedestrian~\cite{zhang2019widerperson}}}
\newcommand{\crowdhuman}[1]{{CrowdHuman~\cite{shao2018crowdhuman}}}
\newcommand{\mobnet}[1]{{MobileNet~\cite{howard2017mobilenets}}}

\newcommand{\caltecha}[1]{{Caltech}}
\newcommand{\citypersona}[1]{{CityPersons}}
\newcommand{\ecpa}[1]{{ECP}}
\newcommand{\widerpersona}[1]{{Wider Pedestrian}}
\newcommand{\crowdhumana}[1]{{CrowdHuman}}
\newcommand{\ours}[1]{{Cascade R-CNN\textsuperscript{$\dagger$}}}

\ifCLASSINFOpdf
\else
\fi

\begin{document}
%
\title{Pedestrian Detection: Domain Generalization, CNNs, Transformers and Beyond}
%
%
%
%

        
\author{Irtiza~Hasan,
        Shengcai Liao,~\IEEEmembership{Senior Member,~IEEE,}
        Jinpeng Li,
        Saad Ullah Akram,
        and~Ling~Shao,~\IEEEmembership{Fellow,~IEEE,}

\IEEEcompsocitemizethanks{\IEEEcompsocthanksitem Irtiza Hasan is with Group 42, UAE. Part of this work was done while he was a researcher at Inception Institute of Artificial Intelligence. \protect\\
irtiza.hasan@inceptioniai.org
\IEEEcompsocthanksitem Shengcai Liao is with Inception Institute of artificial Intelligence, UAE. He is also the corresponding author.\\
shengcai.liao@inceptioniai.org
\IEEEcompsocthanksitem Jinpeng Li is with Department of Computer Science and Engineering, The Chinese University of Hong Kong.
\IEEEcompsocthanksitem Saad Ullah Akram is with Aalto Univeristy, Finland.
\IEEEcompsocthanksitem Ling Shao is with National Center for Artificial Intelligence, Riyadh, Saudi Arabia.

}
\thanks{}}

%
%

\markboth{ieee transactions on pattern analysis and machine intelligence}%
{Shell \MakeLowercase{\textit{et al.}}: Bare Advanced Demo of IEEEtran.cls for IEEE Computer Society Journals}
%



\IEEEtitleabstractindextext{%
\begin{abstract}
Pedestrian detection is the cornerstone of many vision based applications, starting from object tracking to video surveillance and more recently, autonomous driving. 
With the rapid development of deep learning in object detection, pedestrian detection has achieved very good performance in traditional single-dataset training and evaluation setting. 
However, in this study on generalizable pedestrian detectors, we show that, current pedestrian detectors poorly handle even small domain shifts in cross-dataset evaluation. 
We attribute the limited generalization to two main factors, the method and the current sources of data. 
Regarding the method, we illustrate that biasness present in the design choices (e.g anchor settings) of current pedestrian detectors are the main contributing factor to the limited generalization. 
Most modern pedestrian detectors are tailored towards target dataset, where they do achieve high performance in traditional single training and testing pipeline, but suffer a degrade in performance when evaluated through cross-dataset evaluation. Consequently, a general object detector performs better in cross-dataset evaluation compared with state of the art pedestrian detectors, due to its generic design. 
As for the data, we show that the autonomous driving benchmarks are monotonous in nature, that is, they are not diverse in scenarios and dense in pedestrians. Therefore, benchmarks curated by crawling the web (which contain diverse and dense scenarios), are an efficient source of pre-training for providing a more robust representation.
Accordingly, we propose a progressive fine-tuning strategy which improves generalization.  
Additionally, this work also investigate the recent Transformer Networks as backbones to test generalization. 
We demonstrate that as of now, CNNs outperform transformer networks in terms of generalization and absorbing large scale datasets for learning robust representation. 
In conclusion, this paper suggests a paradigm shift towards cross-dataset evaluation, for the next generation of pedestrian detectors. Code and models can be accessed at \url{https://github.com/hasanirtiza/Pedestron}.

\end{abstract}

}

\maketitle

\IEEEdisplaynontitleabstractindextext


%
\IEEEpeerreviewmaketitle


\section{Introduction}
\label{sec:intro}

\begin{figure*}[]
	\begin{center}
	    \includegraphics[width=1\linewidth]{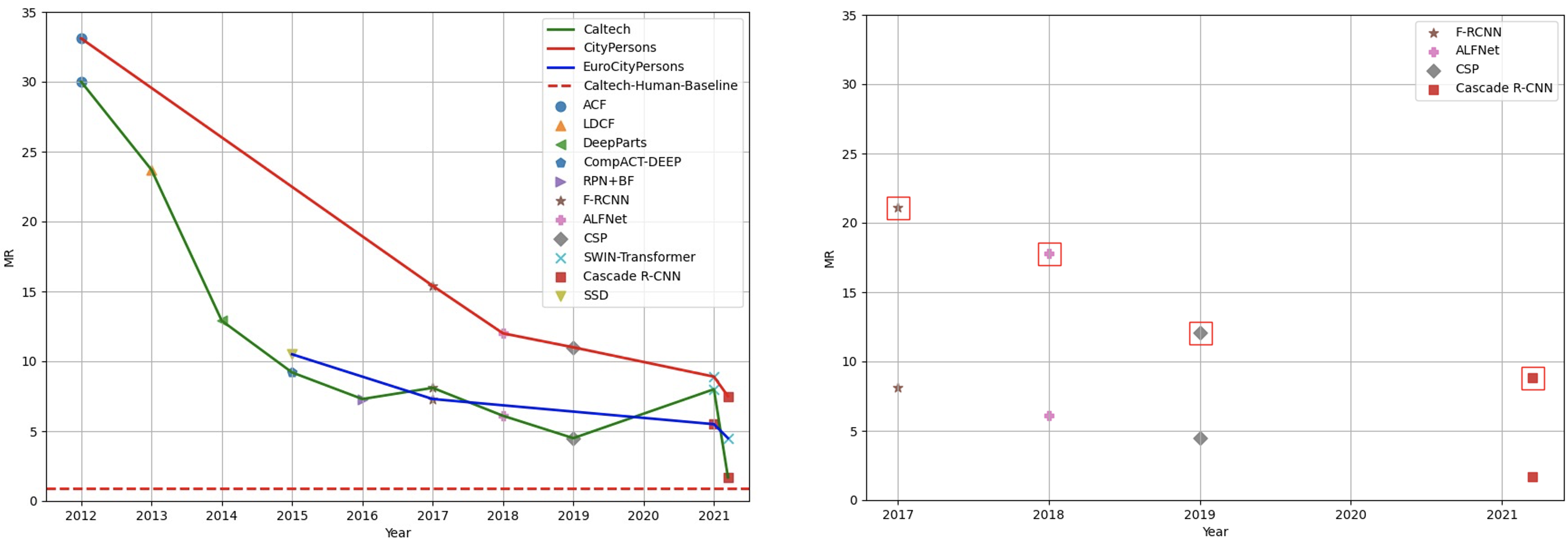}
	\end{center}
\caption{Left: 
Pedestrian detection performance over the years for \textit{Caltech}, \textit{CityPersons} and \textit{EuroCityPersons} on the reasonable subset.
\textit{EuroCityPersons} was released in 2018 but we include results of few older models on it as well.
Dotted line marks the human performance on \textit{Caltech}. Right: We show comparison between traditional single-dataset train and test evaluation on \caltech{} \emph{vs.} cross-dataset evaluation for three pedestrian detectors and one general object detector (Cascade R-CNN). Methods enclosed with bounding boxes are trained on \cityperson{} and evaluated on \caltech{}, while others are trained on \caltecha{}. }
\label{fig:progress}
\end{figure*}

Pedestrian detection is a critical component of many real-world applications, such as, autonomous driving \cite{campmany2016gpu,hbaieb2019pedestrian}, robotic navigation, and video surveillance \cite{hattori2015learning}.
It is one of the most heavily researched computer vision problems and has seen huge improvement over the years.
In last decade, like most other computer vision problems, deep learning based techniques have enabled significant progress for the pedestrian detection.
Currently, on certain benchmarks, the automated methods are approaching the human performance as shown in Fig. \ref{fig:progress} left.
However, current methods are not very robust and their performance varies significantly across the datasets as can be seen in Table~\ref{tab:icon-ch-wider}.
Some widely-used pedestrian detection methods suffer from over-fitting to source datasets, especially in the case of autonomous driving, as shown in Fig. \ref{fig:progress} right.
The performance of these detectors on a target dataset is considerably worse, even when trained on a relatively large source dataset, without any obvious domain shift.
This lack of robustness limits the full potential of these methods as most applications utilizing pedestrian detectors require very low failure rate.

So far, robustness and generalizability, despite being vital for most applications, have not been the main focus in the pedestrian detection research.
Performance characteristics of pedestrian detectors for the same source and target dataset have previously been investigated, but poor cross-dataset performance has not been thoroughly investigated or discussed.
In this paper, we hypothesize that the poor cross-dataset performance is primarily due to over-fitting caused by the fact that the current state-of-the-art pedestrian detectors are heavily tailored for target datasets, which biases their overall design towards the target datasets.

In addition, we hypothesize that the training datasets are generally not dense in pedestrians, lack diversity of environments and scenerios, and have relatively small number of pedestrians, which also limits the performance of current pedestrian detectors.
Almost all state-of-the-art pedestrian detection methods are based on deep learning and their performance depend heavily on the quantity and quality of data, and there is evidence that the performance on some other computer vision tasks (e.g. image classification, object detection, image segmentation) keeps improving at least up-to billions of samples \cite{mahajan2018exploring}.
All current pedestrian detection datasets for autonomous driving have at least three main limitations. 
Firstly, they contain limited number of unique pedestrians.
Secondly, they have low pedestrian density, i.e. relatively small proportion of pedestrians have challenging occlusions. 
The performance of current method is significantly worse for these challenging occluded pedestrians as can be seen in Table~\ref{tab:all-evalbench}.
Thirdly, these datasets have limited diversity as they are captured during small number of sessions by a small team, primarily for dataset creation. 
Thesedays, dashcam videos are widely available online, e.g. youtube, facebook, etc, enabling the potential curation of much more diverse and realistic datasets.

In recent years, few large and diverse person detection datasets, e.g. CrowdHuman \cite{shao2018crowdhuman}, WiderPerson~\cite{zhang2019widerpersondataset} and \widerpersona{} \cite{zhang2019widerperson}, have been created using images and videos available online and through surveillance cameras.
These datasets advance the general person detection research significantly but are not the most suitable datasets for pedestrian detection, as they contain people in a lot more diverse scenarios, than are relevant for autonomous driving.
Nevertheless, they are still beneficial for learning a more general and robust model of pedestrians as they contain more people per image, and they are likely to contain more human poses, appearances and occlusion scenarios, which is beneficial for autonomous driving scenarios, provided current pedestrian detectors have the capacity to digest these large-scale data.

In this paper, we investigate the performance characteristics of current pedestrian detection methods in cross-dataset setting.
We show that 1) the existing methods fare poorly compared to general object detectors, without any adaptations, when provided with larger and more diverse datasets, and 2) when carefully trained, the state-of-the-art general object detectors, without any pedestrian-specific adaptation on the target data, can significantly out-perform pedestrian-specific detection methods on pedestrian detection task (see Fig. \ref{fig:progress} right).
In addition, we propose a progressive training pipeline for better utilization of general person datasets for improving the pedestrian detection performance in autonomous driving scenarios.
We show that by progressively fine-tuning the models, from the dataset furthest from the target domain to the dataset closest to the target domain, large gains in performance can be achieved in terms of \LMR{} on reasonable subset of Caltech (3.7$\%$) and CityPerson (1.5$\%$) without fine-tuning on the target domain.
These improvement hold true for models from all pedestrian detection families that we tested such as Cascade R-CNN \cite{cai2019cascade}, Faster RCNN \cite{ren2015faster} and embedded vision based backbones such as \mobnet . Finally, we also compare the generaliztaion ability of CNNs against the recent transformer network (Swin-Transformer) \cite{liu2021swin}. We illustrate that, despite superior performance of Swin Transformer \cite{liu2021swin}, it struggles when the domain shift is large, in comparison with CNNs. To the best of our knowledge, this is the first study to objectively illustrate this. 

The paper is organized in the following way. Section \ref{sec:prev} reviews the relevant literature. 
We introduce datasets and evaluation protocol in Sec. \ref{sec:experi}. We benchmark our baseline in Sec. \ref{sec:bench}. 
We test the generalization capabilities of the pedestrian specific and general object detectors in Sec. \ref{sec:general-cap}, along with qualitative results. Subsequently, we compare CNNs with Transformer networks in Sec. \ref{sec:cnn_v_trans}. We also discuss effect of fine-tuning on the target set in Sec. \ref{sec:ft}.  
Finally, conclude the paper in Section \ref{sec:conc}.

\section{Related Work} \label{sec:prev}

\noindent\textbf{Pedestrian detection.} Prior to CNNs, the pioneering work of Viola and Jones \cite{viola2004robust} which slid windows over all scales and locations motivated many pedestrian detection methods. To better describe the features of pedestrians, Histogram of Oriented Gradients (HOG) was presented in the work of Dalal and Triggs  \cite{dalal2005histograms}. The aggregate channel feature (ACF) leveraged features in extended channels to improve the speed of pedestrian detection \cite{dollar2014fast}. In similar ways, pedestrian detectors in  \cite{zhang2017citypersons,paisitkriangkrai2014strengthening} employed spatial pooling with low-level features and filtered channel features, respectively. Nonetheless, their performance and generalization ability were still limited by the hand-crafted features.

As the great progress of Convolutional Neural Networks (CNNS), they dominated the research field of generic object detection \cite{ren2015faster,he2017mask,sun2018fishnet,liu2016ssd} and considerably improved the accuracy. The pedestrian detectors \cite{angelova2015real,hosang2015taking,cai2015learning} also benefit from this powerful paradigm. R-CNN detector \cite{girshick2014rich} was utilized in some of the pioneering efforts for pedestrian detection using CNNs \cite{hosang2015taking,zhang2016far} and is still widely employed in this research field. RPN+BF \cite{zhang2016faster} combined Region Proposal Network and boosted forest to enhance the performance of pedestrian detection, which overcame the problems of poor resolution and imbalanced classes in Faster RCNN \cite{ren2015faster}. Although the performance of RPN+BF was outstanding, its learning ability was limited by the non-end-to-end trainable architecture. Since the strong performance and high  expandability of Faster RCNN \cite{ren2015faster} , it inspired a broad spectrum of pedestrian detectors \cite{zhou2018bi,zhang2017citypersons,cai2016unified,brazil2017illuminating,mao2017can}.

Some pedestrian detection methods designed more sophisticated architectures and leveraged extra information to further boost the detection performance. ALF  \cite{liu2018learning} employed several progressive detection heads on Single Shot MultiBox Detector (SSD) \cite{liu2016ssd} to gradually refine the initial anchors, which inherited the merit of high efficiency in single stage detectors and further improved the detection accuracy. Inspired by the blob detection, CSP \cite{Liu2018DBC} reformulated the pedestrian detection task into an anchor-free manner which only needed to locate the center points and regress the scales of pedestrians without relying on the complicated settings of anchor boxes. To improve the detection performance on occluded pedestrians, extra information of the visible-area bounding-box was utilized as the guidance of attention mask in MGAN \cite{pang2019maskguided}. 

\noindent\textbf{Pedestrian detection benchmarks.}
Due to the large practical value of pedestrian detection, a lot of works were devoted to create benchmarks to promote the development of pedestrian detection, such as Daimler-DB \cite{munder2006experimental}, TownCenter \cite{benfold2011stable}, USC \cite{wu2007cluster}, INRIA \cite{dalal2005histograms}, ETH \cite{ess2007depth},and  TUDBrussels \cite{wojek2009multi}, which were all from surveillance scenarios and not suitable for applications in autonomous driving. Recently, the great progress of pedestrian detection also attracted the attention of autopilot systems, and several datasets were created for this context, such as \caltech{}, KITTI \cite{geiger2012we}, \cityperson{}  and  \ecp{}. The cameras in these datasets were typically installed on the front windshield of cars to collect images from the similar field of views as human drivers. \caltech{} and \cityperson{} are the most popular benchmarks for recent learning-based pedestrian detectors, while their small data sizes and monotonous scenarios limit their capabilities in training more robust methods. To solve these limitations, \ecp{} dataset collected images from diverse scenarios including various cities, all seasons, and day and night times, which contains almost ten times more images and eight times more persons than \cityperson{}. Although \ecp{} has a much larger scale, it still suffers from the low density of persons and high similarity of background scenes, which could be the focus of future datasets.  Thus, in this work, we argue that the low density and diversity of these datasets constrains the generalization ability of pedestrian detectors, while the web crawled datasets, such as \crowdhuman{},  WiderPerson~\cite{zhang2019widerpersondataset} and  \widerperson{}, including much diverse scenes and denser persons may increase the upper bound of pedestrian detectors’ generalization ability.

\noindent\textbf{Cross-dataset evaluation.}
Some existing works \cite{zhang2017citypersons, braun2018eurocity,shao2018crowdhuman} explored the relations between the performance of pedestrian detectors and training datasets, whose purpose was to show how much performance advantage could be obtained on target datasets by pre-training on more diverse and dense scenes datasets. But, in this work, we aim to thoroughly evaluate the generalization abilities of some popular pedestrian detection methods by using the cross-dataset evaluation.

\section{Experiments} \label{sec:experi}
\subsection{Experimental Settings}

\begin{table}[!tbh]
\centering
\caption{Experimental settings.}
\label{tab:exp-setting}
\begin{tabular}{l|c|c}
\hline
Setting             & Height & Visibility \\ \hline
Reasonable    & [50, inf]  &    [0.65, inf] \\ \hline
Small    & [50, 75] & [0.65, inf]    \\ \hline
Heavy &  [50, inf] &    [0.2, 0.65]\\ \hline
Heavy* &  [50, inf] &    [0.0, 0.65]\\ \hline
All &   [20, inf]  & [0.2, inf]  \\ \hline
\end{tabular}
\end{table}

\begin{table*}[!tbh]
\centering
\caption{Datasets statistics. $\ddagger$ Fixed aspect-ratio for bounding boxes.}
\label{tab:data-stat}
\begin{tabular}{l|c| c| c| c| c}
\hline
 & Caltech $\ddagger$ & CityPersons $\ddagger$ & ECP & CrowdHuman& Wider Pedestrian\\ \hline
images &42,782  & 2,975  & 21,795 & 15,000 &90,000  \\ \hline
persons &13,674  &19,238  & 201,323 &339,565  &287,131  \\ \hline
persons/image &0.32  &6.47  & 9.2 &22.64  &3.2  \\ \hline
unique persons &1,273  &19,238  & 201,323 &339,565  & 287,131  \\ \hline
\end{tabular}
\end{table*}

\begin{table*}[!tbh]
\centering
\caption{Evaluating generalization abilities of different backbones using our baseline detector.}
\label{tab:backbone}
\begin{tabular}{l|c|c|c}
\hline
Backbone & Training & Testing &  Reasonable  \\ \hline
HRNet  & WiderPedestrian + CrowdHuman  &CityPersons& \textbf{12.8}   \\ \hline
ResNeXt  & WiderPedestrian + CrowdHuman  &CityPersons& \textbf{12.9}  \\ \hline
Resnet-101  & WiderPedestrian + CrowdHuman  &CityPersons& 15.8   \\ \hline
ResNet-50  & WiderPedestrian + CrowdHuman & CityPersons& 16.0    \\ \hline
\end{tabular}
\end{table*}

\textbf{Datasets.} We conduct extensive experiments on three public pedestrian detection datasets which collectd from the scenario of autonomous driving to evaluate and compare with the state-of-the-art pedestrian detection algorithms. These three benchmarks include Caltech~\cite{dollar2012pedestrian}, CityPersons~\cite{zhang2017citypersons} and EuroCity Persons~\cite{braun2018eurocity}, and are categorized into the \emph{autonomous driving} datasets in this work. Caltech~\cite{dollar2012pedestrian} is one of the most popular dataset in the research field of pedestrian detection. It recorded 10 hours of video in Los Angeles, USA by a front-view camera of vehicle, and contained roughly 43K images and 13K persons extracted from the video. We perform the evaluations on the refined Caltech annotations from~\cite{zhang2016far}. Compared to Caltech, \textbf{CityPersons} \cite{zhang2017citypersons} is built upon the dataset of Cityscapes, and contains more diverse scenarios. It was recorded from the street scenes of different cities in and close to Germany. CityPersons contains 2,975, 500, 1,575 images in the training, validation and testing sets, and provides the full bounding-boxes and visible bounding-boxes for 31k pedestrians. \textbf{EuroCity Persons} (ECP) \cite{braun2018eurocity} is a recently released larege-scale dataset recoded in 31 different European cities, which contains more diverse scenarios and is more challenging for pedestrian detectors compared to the datasets of Caltech and CityPersons. It contains two subsets of ECP day-time and ECP night-time based on the recording time. ECP dataset has roughly 200K bounding-boxes. Similar to the evaluation procedure in \ecp{}, we conduct the experiments on the subset of ECP day-time for the fair comparisons with existing literature. Unless other statement, all experimental results are from the validation sets due to the frequent submissions to online testing server is not allowed. Except to the \emph{\textbf{autonomous driving}} datasets of \caltecha{}, \citypersona{} and \ecpa{}, we further conduct the experiments on two \emph{\textbf{web-crawled}} datasets of \crowdhumana{} and \widerpersona{}\footnote{\widerpersona{} contains images from the scenarios of autonomous driving and surveillance. The data provided in 2019 challenge was used in our experiments. Data can be downloaded from : \url{https://competitions.codalab.org/competitions/20132}} \cite{zhang2019widerperson}. We provide more details of above datasets in Table \ref{tab:data-stat}.

\textbf{Evaluation protocol.} We evaluate the performance of pedestrian detectors by the widely used metric of log average miss rate over False
Positive Per Image (FPPI) over range [$10^{-2}$, $10^{0}$] (\LMR{}) on \caltech{}, \cityperson{} and \ecp{}. The experimental results on different occlusion levels including \textbf{Reasonable}, \textbf{Small}, \textbf{Heavy}, \textbf{Heavy*}\footnote{In the case of \citypersona{}, for the fair comparison with some previous methods under the same setting, we also report the numbers under the visibility level between [0.0,0.65] which is denoted as \textbf{Heavy*} occlusion.} and \textbf{All} are reported unless stated otherwise. Table \ref{tab:exp-setting} provides the specific settings of each set.   

\textbf{Cross-dataset evaluation.} To evaluate the generalization ability of pedestrian detectors, we perform cross-dataset evaluation by only using the training set of dataset \emph{A} to train models and directly testing then on the validation/testing set of dataset \emph{B}. This training and testing procedure is consistent for all experiments, and denoted as A$\rightarrow$B.

\textbf{Baseline.} Because most of the high performance pedestrian detectors on the datasets of Caltech, CityPersons and ECP are built upon the two-stage detectors of Faster/Mask R-CNN~\cite{ren2015faster,he2017mask}, the more powerful multi-stage detector of \textbf{Cascade R-CNN \cite{cai2019cascade}} which also belongs to the R-CNN family is chosen as the \textbf{baselines}. In this work, the terms of baseline and Cascade RCNN are interchangeably used, which both means the same method of Cascade R-CNN~\cite{cai2019cascade}. Multiple detection heads are used step-by-step in the Cascade R-CNN to gradually filter out the false positive anchors and generate more and more high-quality proposals. We equipped different backbones on our baseline method to evaluate its robustness, and the details of these experiments are shown in Table \ref{tab:backbone}. Among them, ResNeXt \cite{xie2017aggregated} and HRNet \cite{wang2019deep} show the top ranked performance. Unless other statements, HRNet \cite{wang2019deep} is used as the default backbone network of our baseline. HRNet simultaneously processes multiple levels feature maps in a parallel way retaining both low-level details and high-level semantic information, which may greatly benefit the pedestrian detection under the large scale variations.

\section{Benchmarking}
\label{sec:bench}
\vspace{-2mm}

The benchmarking results of our baseline, Cascade R-CNN \cite{cai2019cascade}, on three autonomous driving datasets including \caltech{} dataset, \cityperson{} and on \ecp{} are presented in Table \ref{tab:all-evalbench}. Without any "bells and whistles", our baseline achieved the performance comparable to the specially customized pedestrian detectors on the datasets of Caltech and CityPersons. Interestingly, the performance gap between our baseline and the state-of-the-art algorithms changes as the sizes of datasets increase. The relative performance of our baseline is lowest on the smallest dataset of Caltech and significantly improves on the largest dataset of ECP.

\begin{table}[!tbh]
\centering
 \caption{Benchmarking on autonomous driving datasets.}
  \label{tab:all-evalbench}
  \resizebox{0.99\linewidth}{!}{
\begin{tabular}{l|c|c|c|c}
\hline
Method     & Testing    & Reasonable & Small &Heavy   \\ \hline
ALFNet \cite{liu2018learning} & Caltech  & 6.1     &  7.9 & 51.0   \\ \hline
Rep Loss \cite{wang2018repulsion}&    Caltech    & \textbf{5.0}   & \textbf{5.2}    & 47.9   \\ \hline
CSP \cite{Liu2018DBC}  &Caltech & \textbf{5.0}     &  6.8  & \textbf{46.6}   \\ \hline
Cascade R-CNN \cite{cai2019cascade}    &  Caltech  & 6.2       & 7.4 & 55.3   \\ \hline

\hline \hline
RepLoss \cite{wang2018repulsion} &CityPersons &13.2 & - & - \\
\hline
ALFNet \cite{liu2018learning} & CityPersons&12.0 & 19.0 & 48.1  \\
\hline
CSP \cite{Liu2018DBC} &CityPersons& \textbf{11.0} & 16.0 &39.4 \\
\hline
Cascade R-CNN \cite{cai2019cascade} & CityPersons& {11.2} & \textbf{14.0} & \textbf{37.1} \\
\hline

\hline \hline
Faster R-CNN \cite{braun2018eurocity}   &ECP& 7.3  & 16.6     & 52.0 \\ \hline
YOLOv3 \cite{braun2018eurocity}    &ECP&  8.5 & 17.8     & 37.0\\ \hline
SSD \cite{braun2018eurocity} &  ECP&10.5 & 20.5   & 42.0 \\ \hline
Cascade R-CNN \cite{cai2019cascade} &ECP &\textbf{6.6}    & \textbf{13.6}   & \textbf{33.3} \\ \hline
\end{tabular}
}
\end{table}

\section{Generalization Capabilities} \label{sec:general-cap}

As previously mentioned, existing works evaluate the pedestrian detectors with a traditional manner where training and evaluation data are from the same domain dataset, i.e., within-dataset evaluation. However, we argue that this algorithm development pipeline ignores the generalization capability of pedestrian detectors and is easy to over-fit on a specific dataset. Thus, in this work, we emphasize the importance of cross-dataset evaluation in the design of pedestrian detectors. We can clearly show how well pedestrian detectors perform on unseen domain by cross-dataset evaluation. Thus, extensive cross-dataset experiments are conducted in this section to evaluate the robustness of pedestrian detectors.

\subsection{Dataset Illustrations}

\begin{figure*}[!tbh]
	\begin{center}
	    \includegraphics[width=1\linewidth]{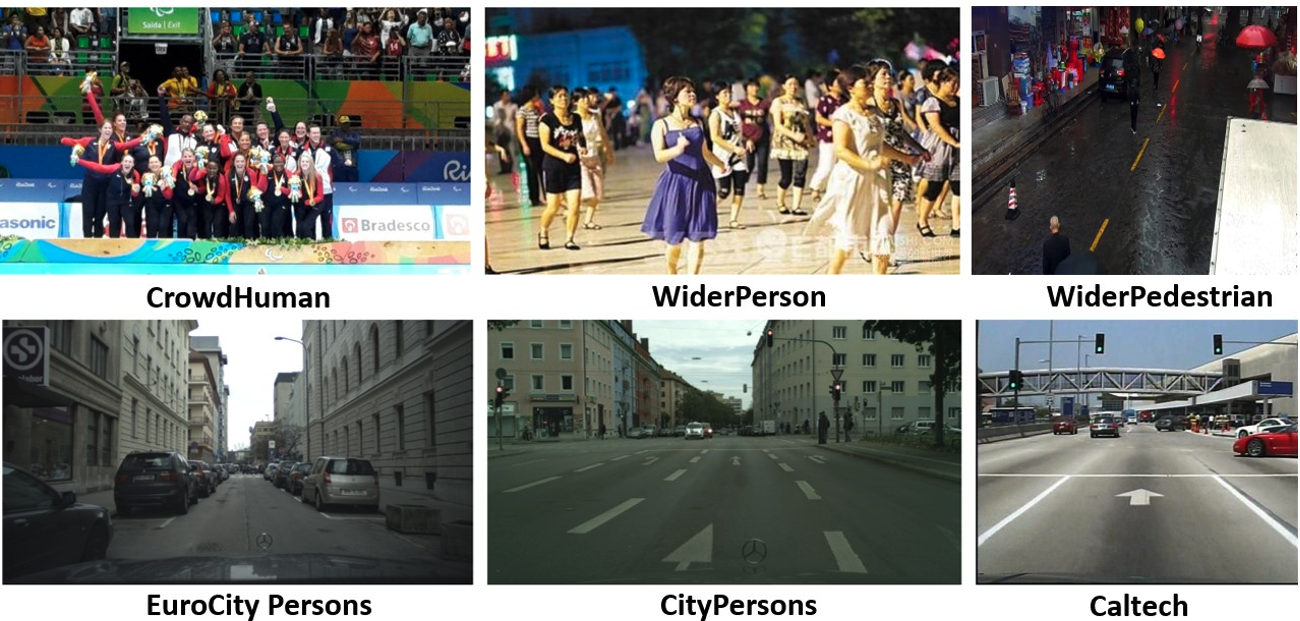}
	\end{center}
	\caption{Illustration of benchmarks. Top row shows images from diverse and dense datasets, such as \crowdhuman{} and \widerperson{}. Bottom row presents images of autonomous driving benchmarks, \ecp{}, \cityperson{} \caltech{}
	}
	\label{fig:dat-exp}
\end{figure*}

We showcase some examples of datasets related to pedestrian detection in Figure \ref{fig:dat-exp}. Top row depicts different scenarios in diverse and dense datasets collected by crawling on web. Bottom row illustrates images from traditional autonomous driving datasets. It can be observed that web-crawled datasets provide more enriched representation of pedestrians, since they cover several scenarios, such as different poses, illumination and different types of occlusion. Whereas, autonomous driving benchmarks are monotonous in nature, i.e. same background, view-point etc. Interestingly, \ecp{} and \cityperson{}, illustrate striking resemblance (where the camera is mounted, image resolution, geographical location etc.), this further stresses the point that even when the target domains are not drastically different, current pedestrian detectors do not generalize well (cf. Tables 5,6 and 7 in the paper).    

\begin{table*}[!tbh]
\centering
\caption{Cross dataset evaluation on Caltech and CityPersons. A$\rightarrow$B refers to training on \emph{A} and testing on \emph{B}.}
\label{tab:cross-cal-cp}
\resizebox{0.99\linewidth}{!}{
\begin{tabular}{l|c|c|c|c|c}
\hline
Method & Bakcbone &CityPersons$\rightarrow$CityPersons  & CityPersons$\rightarrow$Caltech & 
Caltech$\rightarrow$Caltech & Caltech$\rightarrow$CityPersons \\ \hline
FRCNN \cite{zhang2017citypersons} &VGG-16& 15.4  & 21.1 & 8.7 & 46.9  \\ \hline
Vanilla FRCNN \cite{zhang2017citypersons}  &VGG-16& 24.1  & 17.6 & 12.2 & 52.4  \\ \hline
ALFNET \cite{liu2018learning} &ResNet-50& 12.0  & 17.8& 6.1  & 47.3 \\ \hline
CSP \cite{Liu2018DBC} &ResNet-50& 11.0  & 12.1 & 5.0 & 43.7  \\ \hline
PRNet \cite{song2020progressive} &ResNet-50&  10.8 & 10.7& - & - \\ \hline
BGCNet \cite{li2020box} &HRNet & \textbf{8.8} & 10.2 & \textbf{4.1} & 41.4 \\ \hline
Faster R-CNN \cite{ren2015faster} &ResNext-101& 16.4 & 11.8 &  9.7 & 40.8   \\ \hline
Cascade R-CNN \cite{cai2019cascade} &HRNet&  11.2 & \textbf{8.8} & 6.2  & \textbf{36.5} \\ \hline
Cascade R-CNN \cite{cai2019cascade} &Swin Transformer&  9.2 & {9.1} & 8.0  & 41.9 \\ \hline

\end{tabular}
}
\end{table*}


\begin{table*}[!tbh]
\centering
\caption{Cross dataset evaluation of (Casc. R-CNN and CSP) on Autonomous driving benchmarks. Both detectors are trained with HRNet as a backbone.}
\label{tab:cross-cp-ecp-cal}
\begin{tabular}{l|c|c|c|c|c}
\hline
Method & Training   & Testing          & Reasonable & Small & Heavy \\ \hline
Casc. RCNN & CityPersons & CityPersons  & {11.2}  & 14.0     & {37.0} \\ \hline
CSP & CityPersons & CityPersons  & \textbf{9.4}  & \textbf{11.4}     & \textbf{36.7} \\ \hline
Casc. RCNN  & ECP   & CityPersons&  {10.9} & \textbf{11.4}     & 40.9\\ \hline
CSP & ECP   & CityPersons&  {11.5} & {16.6}     & 38.2\\ \hline
\hline 
Casc. RCNN  & ECP    &ECP& \textbf{6.9}  & \textbf{12.6}     & \textbf{33.1} \\ \hline
CSP & ECP    &ECP& {19.4}  & {50.4}     & {57.3} \\ \hline
Casc. RCNN & CityPersons & ECP   &  17.4 & 40.5     & 49.3\\ \hline
CSP & CityPersons & ECP   &  19.6 & 51.0     & 56.4\\ \hline
\hline 
Casc. RCNN & CityPersons & Caltech  & 8.8    &  9.8    & \textbf{28.8} \\ \hline
CSP & CityPersons & Caltech  & 10.1    &  13.3    & {34.4} \\ \hline
Casc. RCNN & ECP & Caltech& \textbf{8.1}   & \textbf{9.6}   & 29.9 \\ \hline
CSP & ECP & Caltech&10.4   & 13.7   & 31.3 \\ \hline
\end{tabular}
\end{table*}

\subsection{Cross Dataset Evaluation of Existing State-of-the-Art} \label{subsec:cross dataset}
In this section we demonstrate that existing state-of-the art pedestrian detectors generalize worse than general object detector. 
We show that this is mainly due to the biases in the design of methods for the target set, even when other factors, such as backbone, are kept consistent.  

To see how well state-of-the-art pedestrian detectors generalize to different datasets, we performed cross dataset evaluation of five state-of-the-art pedestrian detectors and our baseline (Cascade RCNN) on \cityperson{} and \caltech{} datasets. 
We evaluated recently proposed BGCNet \cite{li2020box}, CSP \cite{Liu2018DBC}, PRNet \cite{song2020progressive}, ALFNet \cite{liu2018learning} and FRCNN \cite{zhang2017citypersons}(tailored for pedestrian detection). 
Furthermore, we added along with baseline, \emph{Faster R-CNN \cite{ren2015faster}}, without \say{bells and whistles}, but with a more recent backbone ResNext-101 \cite{xie2017aggregated} with FPN \cite{lin2017feature}. Moreover, we implemented a vanilla \emph{FRCNN \cite{zhang2017citypersons}}  with VGG-16 \cite{simonyan2014very} as a backbone and with no pedestrian specific adaptations proposed in \cite{zhang2017citypersons} (namely quantized anchors, input scaling, finer feature stride, adam solver, ignore region handling, etc).

We present results for \caltecha{} and \citypersona{} in Table \ref{tab:cross-cal-cp}, respectively. 
We also report results when training is done on target dataset for readability purpose. For our results presented in Table \ref{tab:cross-cal-cp} (Fourth column, CityPersons$\rightarrow$Caltech), we trained each detector on \citypersona{} and tested on \caltecha{}. Similarly, in the last column of the Table \ref{tab:cross-cal-cp}, all detectors were trained on the \caltecha{} and evaluated on \citypersona{} benchmark. 
As expected, all methods suffer a performance drop when trained on \citypersona{} and tested on \caltecha{}. Particularly, BCGNet \cite{li2020box}, CSP \cite{Liu2018DBC}, ALFNet \cite{liu2018learning} and FRCNN \cite{zhang2017citypersons} degraded by more than 100 $\%$ (in comparison with fifth column, Caltech$\rightarrow$Caltech). 
Whereas in the case of Cascade R-CNN \cite{cai2019cascade}, performance remained comparable to the model trained and tested on target set. 
Since, \citypersona{} is a relatively diverse and dense dataset in comparison with \caltecha{}, this performance deterioration cannot be linked to dataset scale and crowd density. 
This illustrates better generalization ability of general object detectors over state-of-the-art pedestrian detectors. Moreover, it is noteworthy that BGCNet \cite{li2020box} like the Cascade R-CNN \cite{cai2019cascade}, also uses HRNet \cite{wang2019deep} as a backbone, making it directly comparably to the Cascade R-CNN \cite{cai2019cascade}.  

Importantly, pedestrian specific FRCNN \cite{zhang2017citypersons} performs worse in cross dataset (fourth column only), compared with its direct variant vanilla FRCNN. The only difference between between the two being pedestrian specific adaptations for the target set, highlighting the bias in the design of tailored pedestrian detectors. 

Similarly, standard Faster R-CNN \cite{ren2015faster}, though performs worse than FRCNN \cite{zhang2017citypersons} when trained and tested on the target dataset, it performs better than FRCNN \cite{zhang2017citypersons} when it is evaluated on \caltecha{} without any training on \caltecha{}.

It is noteworthy that Faster R-CNN \cite{ren2015faster} outperforms state-of-the-art pedestrian detectors (except for BGCNet \cite{li2020box}) as well in cross dataset evaluation, presented in Table \ref{tab:cross-cal-cp}. 
We again attribute this to the \emph{bias} present in the design of current state-of-the-art pedestrian detectors, which are tailored for specific datasets and therefore limit their generalization ability. 
Moreover, a significant performance drop for all methods (though ranking is preserved except for vanilla FRCNN), including Cascade R-CNN \cite{cai2019cascade}, can be seen in Table \ref{tab:cross-cal-cp}, last column. However, this performance drop is attributed to lack of diversity and density of the \caltecha{} dataset. \caltecha{} dataset has less annotations than \citypersona{} and number of people per frame is less than 1 as reported in Table \ref{tab:data-stat}. However, still it is important to highlight, even when trained on a limited dataset, usually general object detectors are better at generalization than state-of-the-art pedestrian detectors. Interestingly, Faster R-CNN's \cite{ren2015faster} error is nearly twice as high as that of BGCNet \cite{li2020box} in within-dataset evaluation, whereas it outperforms in BGCNet \cite{li2020box} in cross-dataset evaluation.

As discussed previously, most pedestrian detection methods are extensions of general object detectors (FR-CNN, SSD, etc.). However, they adapt to the task of pedestrian detection. 
These adaptations are often too specific to the dataset or detector/backbones (e.g. anchor settings \cite{zhang2017citypersons, liu2018learning}, finer stride \cite{zhang2017citypersons}, additional annotations \cite{zhou2018bi,pang2019maskguided}, constraining aspect-ratios and fixed body-line annotation \cite{Liu2018DBC,li2020box} etc.). These adaptations usually limit the generalization as shown in Table \ref{tab:cross-cal-cp}, also discussed, task specific configurations of anchors limits generalization as discussed in \cite{liu2019center}.

\subsection{Autonomous Driving Datasets for Generalization} \label{gen:ad}

We show that the general object detectors outperforms existing pedestrian detection methods (such as CSP~\cite{Liu2018DBC}) as learning a generic feature representation for pedestrians, even when they are trained on the large dataset (such as ECP) and tested on the small dataset (such as Caltech). Furthermore, detectors achieve higher generalization from larger and denser autonomous driving datasets.

As shown in the last section, cross dataset evaluation can shed light on the generalization capabilities of pedestrian detectors. Moreover, the characteristic of dataset is also an important determinant for the model generalization. The intrinsic nature of the real world can be more effectively draw by diverse datasets~\cite{braun2018eurocity}. Consequently, such datasets potentially provide the chances for pedestrian detectors to learn more generic feature representation to robustly tackle the domain shifts. Instead of exploring the impact of dataset in generalization as existing studies ~\cite{braun2018eurocity,shao2018crowdhuman,zhang2019widerpersondataset}, we aim at presenting a detailed comparison of a general object detector and state-of-the-art pedestrian detection methods when the training and testing datasets are varied. For fair comparsions, the backbone of CSP~\cite{Liu2018DBC} is replaced from ResNet-50 to HRNet~\cite{wang2019deep}. As shown in the second row of Table~\ref{tab:cross-cp-ecp-cal}, this change improves the performance of CSP from 11.0$\%$\LMR ~~to 9.4$\%$\LMR.



First, we train Cascade RCNN and CSP on the pedestrian detection dataset of \ecpa{} which contains more countries and cities and is the largest benchmark with regard to pedestrian density and diversity from the context of autonomous driving. \cityperson{} is chosen as the evaluation benchmark, and results are shown in the third and fourth row of Table \ref{tab:cross-cp-ecp-cal}, respectively. It is clear that in the case of providing the same backbone, Cascade RCNN generalizes better on \citypersona{} than CSP in the reasonable setting. Considering CSP significantly outperforms Cascade RCNN by nearly 2$\%$ \LMR ~~points when they are evaluated in the within-dataset setting, it is surprising to see that the results are turn over.


Secondly, we train CSP and Cascade RCNN on \citypersona{ } and evaluate them on \ecp{} to further study the generalization abilities of them under different diverse degrees of training datasets. Similarly, when training dataset is with low diversity, Cascade RCNN still outperforms than CSP. The performance difference is 10.5 $\%$ \LMR~, 7.1 $\%$ \LMR~ and 2.2 $\%$ \LMR~ in the small, heavy and reasonable settings, respectively. 

Finally, we combine \citypersona{} and \ecpa{} as the training data and perform the evaluate on the \caltecha{} which is the smallest data source. The results of Cascade RCNN and CSP in all settings are shown in the last four rows of Table \ref{tab:cross-cp-ecp-cal}. We conclude that when we use a diverse and dense training dataset, \ecpa, Cascade RCNN has more robust performance than CSP on all evaluating subsets.

\begin{table*}[!tbh]
\centering
\caption{Benchmarking with CrowdHuman and Wider Pedestrian dataset.}
\label{tab:icon-ch-wider}
\begin{tabular}{l|c|c|c|c|c}
\hline
Method & Training & Testing             & Reasonable & Small & Heavy \\ \hline
Casc. RCNN & \crowdhumana{}& \caltecha{}    & 3.4  &   11.2   & 32.3 \\ \hline
CSP & \crowdhumana{}& \caltecha{}    &  4.8 &  5.7    & 31.9 \\ \hline
Casc. RCNN& \crowdhumana{}& \citypersona{}    & 15.1   & 21.4      & 49.8 \\ \hline
CSP & \crowdhumana{}& \citypersona{}    & 11.8   & 18.3      & 44.8 \\ \hline
Casc. RCNN & \crowdhumana{}& \ecpa{} & 17.9  & 36.5   & 56.9  \\ \hline

CSP & \crowdhumana{}& \ecpa{} & 19.8  & 48.9   & 60.1  \\ \hline
\hline
Casc. RCNN & \widerpersona{}&\caltecha{}    & 3.2  &   10.8   & 31.7 \\ \hline
CSP & \widerpersona{}&\caltecha{}    & 3.4  &    3.0  & 29.5 \\ \hline
Casc. RCNN& \widerpersona{}&\citypersona{}    & 16.0   & 21.6      & 57.4 \\ \hline
CSP & \widerpersona{}&\citypersona{}    & 17.0   & 22.4 & 58.2 \\ \hline
Casc. RCNN & \widerpersona{}&\ecpa{} & 16.1  & 32.8   & 58.0  \\ \hline
CSP & \widerpersona{}&\ecpa{} &  24.1 & 62.6   & 76.7  \\ \hline
\end{tabular}
\end{table*}

\begin{table*}[!tbh]
\centering
\caption{Investigating the effect on performance when CrowdHuman, Wider Pedestrian and ECP are merged and Cascade R-CNN \cite{cai2019cascade} is trained only on the merged dataset.}
\label{tab:icon-collapsed}
\begin{tabular}{l|c|c|c|c|c}
\hline
Method & Training        & Testing     & Reasonable & Small & Heavy \\ \hline
Casc. RCNN&CrowdHuman $\rightarrow$ ECP & CP & {10.3}     & {12.6}   & {40.7}  \\ \hline
CSP&CrowdHuman $\rightarrow$ ECP & CP & {10.4}     & {10.0}   & {36.2}  \\ \hline
Casc. RCNN&Wider Pedestrian $\rightarrow$ ECP & CP & \textbf{9.7}     & \textbf{11.8}  & {37.7}  \\ \hline
CSP&Wider Pedestrian $\rightarrow$ ECP & CP & {9.8}     & {14.6}  & \textbf{35.4}  \\ 
\hline \hline 
Casc. RCNN&CrowdHuman $\rightarrow$ ECP & Caltech & {2.9}     & {11.4}   & {30.8}  \\ \hline
CSP&CrowdHuman $\rightarrow$ ECP & Caltech & {11.0}     & {14.7}   & {32.2}  \\ \hline
Casc. RCNN &Wider Pedestrian $\rightarrow$ ECP & Caltech & \textbf{2.5}     & \textbf{9.9}   & \textbf{31.0}  \\ \hline
CSP &Wider Pedestrian $\rightarrow$ ECP & Caltech & {8.6}     & {12.0}   & {30.3}  \\ \hline
\end{tabular}
\end{table*}


\subsection{Diverse General Person Detection Datasets for Generalization} \label{gen:div}

We study how much performance improvement dense and diverse datasets can bring. When the testing source is small datasets from the context of autonomous driving, such as \caltech{}, diverse and dense datasets are still beneficial for generalization even under large domain gaps between training and evaluation datasets. Moreover, diverse and dense datasets can bring more benefits to the general object detection methods, such as Cascade RCNN than the specially tailored pedestrian detectors, such as CSP.

\crowdhuman{} and \widerperson{} datasets are two diverse and dense pedestrian detection datasets collected from web-crawling and surveillance cameras.
Unlike the autonomous driving datasets, the crowd density and scenario diversities are large in \crowdhuman{} and \widerperson{} since they include images in diverse sources, such as street views and surveillance, which increases the data diversity from a different form. Thus, they are idea sources to pre-train models. We pre-train Cascade R-CNN~\cite{cai2019cascade} and CSP~\cite{Liu2018DBC} on \crowdhuman{} and \widerperson{} datasets, and shows corresponding results in Table \ref{tab:icon-ch-wider}. 
It can be seen that pre-training significantly boost the performance of pedestrian detectors. When tested on \caltecha{} dataset, Cascade R-CNN outperforms all previous methods which are only trained on \caltecha{}, and the test error is reduced nearly by half. The trend of performance improvement is observed on the results of CSP~\cite{Liu2018DBC}, although its improvement is less than Cascaded R-CNN. 
The performance of either Cascade RCNN or CSP are not improved on \cityperson{}, when they are trained on \crowdhuman{}. This is reasonable due to \cityperson{} is more difficult than \caltech{} with regard to density and diversity. Similar trends are observed in Table \ref{tab:cross-cp-ecp-cal} when detectors are tested on \ecp{}. Training on \cityperson{} brings better performance than training on \crowdhuman{}.    
From the bottom half of Table \ref{tab:icon-ch-wider}, we can see that general object detector benefits more from training on \widerperson{}. We hypothesis this is because that \widerperson{} is with larger scale and more similar to target domain than \crowdhuman{}. The domain difference is reflected in the scenarios of images where \crowdhuman{} includes web-crawled persons with diverse poses while \widerperson{} are mainly from street views and surveillance cameras.

\subsection{Progressive Training Pipeline}


We propose a progressive training pipeline to take full charge of multi-source datasets and thus further improve the pedestrian detection performance. This pipeline first trains detectors on a dataset that is farther from target domain but general diverse enough and then fine-tune them on a dataset which is similar to the target domain.

Extensive experiments are conducted to demonstrate the value of progressive training. To be in line with the study described in the previous section, target domain dataset is not touched, and only the training subset of each corresponding dataset is used in our pipeline. We use the symbol of \emph{A} $\rightarrow$ \emph{B} to denote pre-training model on \emph{A} dataset and fine-tuning it on \emph{B} dataset. Besides, two datasets of \emph{A} and \emph{B} could be directly merged together to train the model, which is denoted as \emph{A} + \emph{B}. In this section, \caltech{} and \cityperson{} datasets are respectively used as the evaluation benchmark, and corresponding results are shown in Table \ref{tab:icon-collapsed}. 
The upper part of Table \ref{tab:icon-collapsed} clearly shows that the performance of Cascade RCNN can be significantly improved by the progressive training pipeline. Noticeably, without training on \cityperson{} dataset, Cascade R-CNN achieves comparable results as the state-of-the-art detectors through the progressive training pipeline of \widerperson{} $\rightarrow$ \ecp{}. Besides, progressive training also helps Cascade R-CNN achieve new state-of-the-art results on the \caltech{} dataset. It is worth noting that our performance on \caltech{} is very close to the human baseline (0.88).

Finally, we show the experimental results of directly merging all datasets in the third and fourth rows of Table \ref{tab:icon-collapsed}. This training strategy can also improve the performance but it still cannot reach the performance of progressive training pipeline, which demonstrates the value of pre-training on general dataset and then fine-tuning on the autonomous driving dataset. Without touching the data in target domain, our progressive training helps to effectively improve the pedestrian detection performance of state-of-the-art detectors. These experiments demonstrate that our training pipeline explores a way to significantly improve the generalization capability of Cascade R-CNN and makes it on a level with state-of-the-art detectors on \cityperson{} and achieve best performance on \caltech{}.

\subsection{Application Oriented Models}
In this section, we conducted experiments to show that pre-training on dense and diverse datasets can help a light-weight neural network architecture, \mobnet{}, to achieve competitive results as state-of-the-art detectors, such as CSP, on \cityperson{} dataset.

The computational cost and model size of pedestrian detectors are important factors in many real-world applications, such as drones and autonomous driving cars, which require real-time detection and are usually run on limited hardware. To study whether progressive training pipeline is still effective in improving the performance of light-weight backbone, we conduct experiments with a widely used light-weight neural network backbone, MobileNet~\cite{howard2017mobilenets} v2, proposed for embedded and mobile computer vision tasks.

We replace the backbone of Cascade R-CNN \cite{cai2019cascade} as a \mobnet{}, and present the results on \cityperson{} in Table \ref{tab:mobileNet}. To establish reference performance, we train and test \mobnet{} on \cityperson{}, and show its results in the first row of Table \ref{tab:mobileNet}. Intuitively, the performance of \mobnet{} is lower than HRNet \cite{wang2019deep}. But, the results demonstrate that progressive training by pre-training \mobnet{} on \crowdhuman{} and then fine-tuning on \ecp{} still effectively improves the detection performance on \cityperson{}.
Moreover, further performance improvement can be achieved by replacing the pre-training dataset of \crowdhuman{} as \widerperson{}. As shown in the first and forth rows, our progressive pipeline of \widerperson{} $\rightarrow$ \ecp{} improves the performance of 0.6$\%$ \LMR on the reasonable subset of \cityperson{}.
The results in Table \ref{tab:mobileNet} and \ref{tab:icon-ch-wider} both demonstrate that pre-training on \widerperson{} dataset can bring larger performance improvement than \crowdhuman{} when the evaluation benchmark is \cityperson{}. This is because \widerperson{} includes scenario of autonomous driving and shares more common characteristics with target domain.
It is worth noting that our progressive training pipeline makes Cascade R-CNN with a light-weight backbone of \mobnet{} approach the performance of state-of-the-art method, CSP~\cite{Liu2018DBC}, which is equipped with a much larger backbone of ResNet-50. 


\begin{table}[!tbh]
\centering
\caption{Investigating  the performance of embedded vision model, when pre-trained on diverse and dense datasets.}
\label{tab:mobileNet}
 \resizebox{0.99\linewidth}{!}{
\begin{tabular}{l|c|c|c|c}
\hline
Training     & Testing        & Reasonable & Small & Heavy \\ \hline
CP   & CP & 12.0     & 15.3     & 47.8 \\ \hline \hline
ECP&  CP &    19.1 & 19.3     &  51.3\\ \hline
CrowdHuman$\rightarrow$ECP&  CP &    11.9 & 15.7     &  48.9\\ \hline
Wider Pedestrian$\rightarrow$ECP  & CP & \textbf{11.4}    & \textbf{14.6}    & \textbf{43.4} \\ \hline
\end{tabular}
}
\end{table}

\subsection{Qualitative Results}
 \begin{figure*}[!tbh]
	\begin{center}
	    \includegraphics[width=1\linewidth]{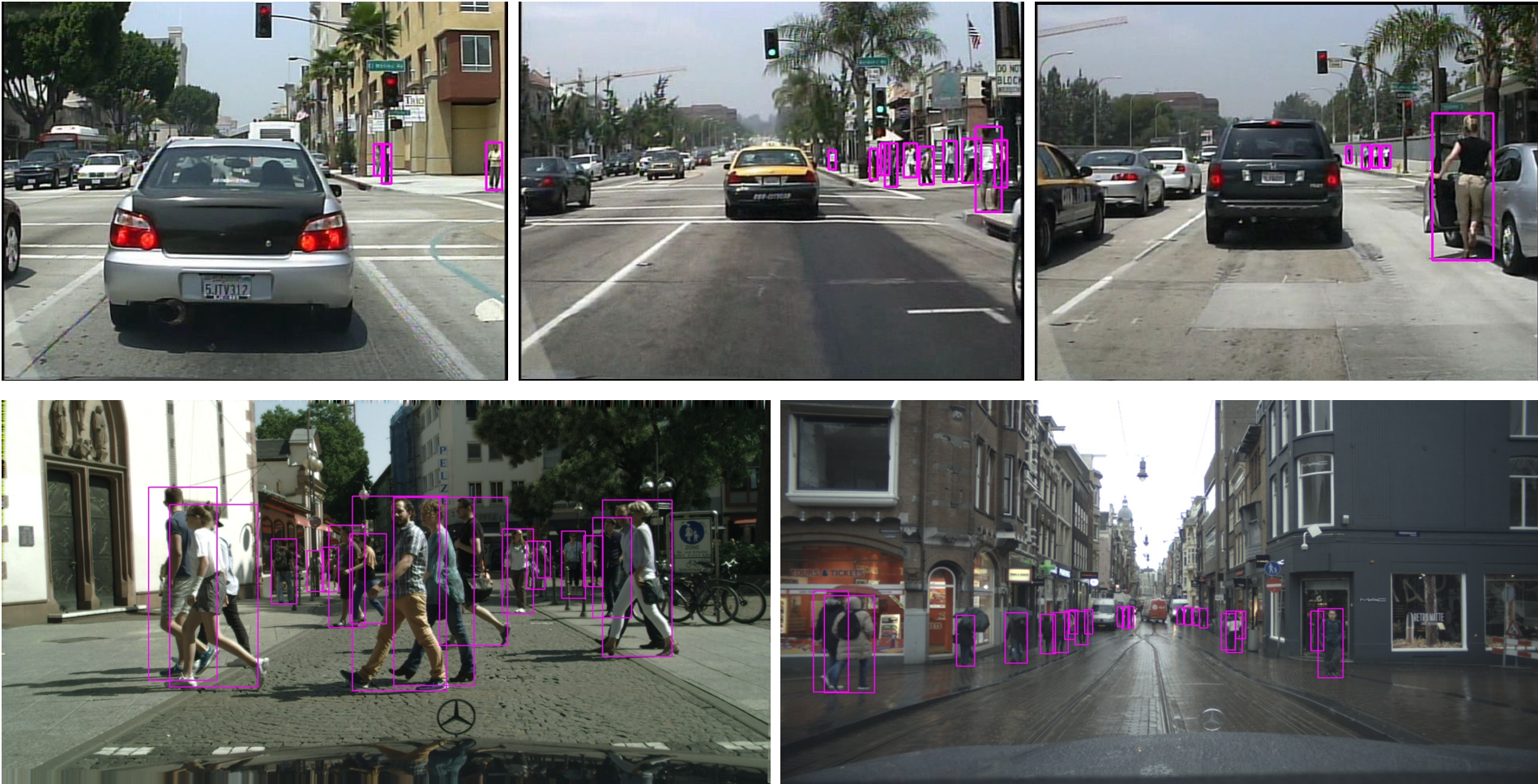}
	\end{center}
	\caption{Qualitative results of Cascade R-CNN on different benchmarks. Top row includes example from Caltech, bottom left CityPersons and ECP on bottom right. \textsuperscript{$\dagger$} }
	\label{fig:good-1}
\end{figure*}

 \begin{figure*}[!tbh]
	\begin{center}
	    \includegraphics[width=1\linewidth]{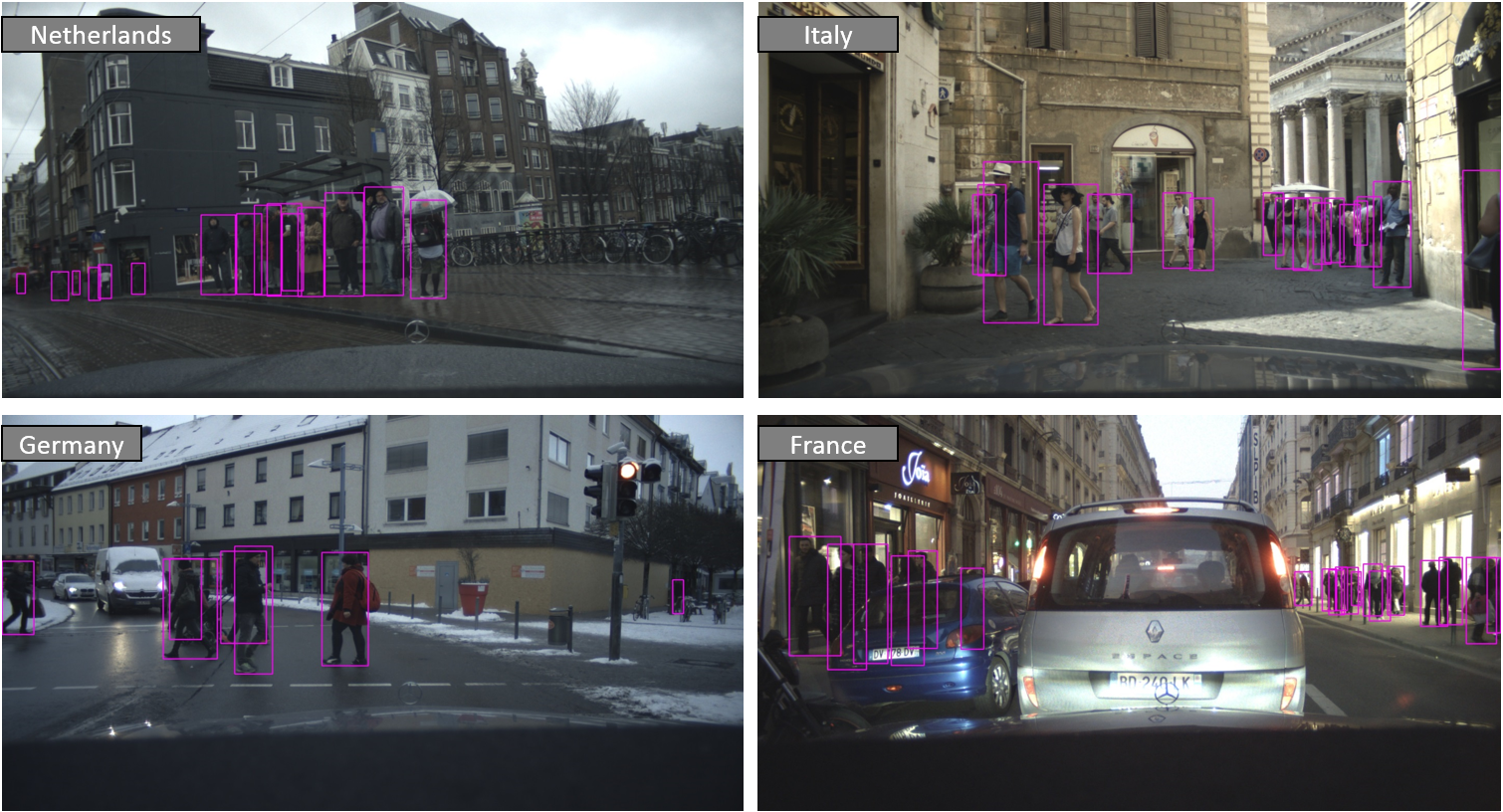}
	\end{center}
	\caption{Cascade R-CNN \textsuperscript{$\dagger$} across different scenarios, such as summer, winter, rain and low-illumination, illustrating the robustness of the general object detector. }
	\label{fig:good-2}
\end{figure*}

\begin{figure*}[!tbh]
\begin{subfigure}{0.33\textwidth}
  \centering
  \includegraphics[height=0.15\textheight]{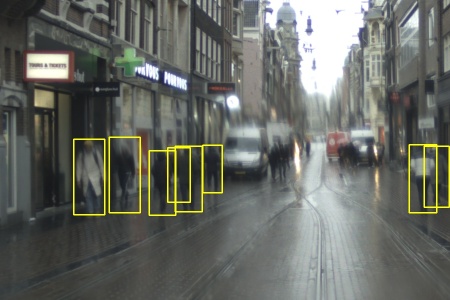}
  \caption{}
  \label{fig:failures:fp:post1}
\end{subfigure}
\begin{subfigure}{0.33\textwidth}
  \centering
  \includegraphics[height=0.15\textheight]{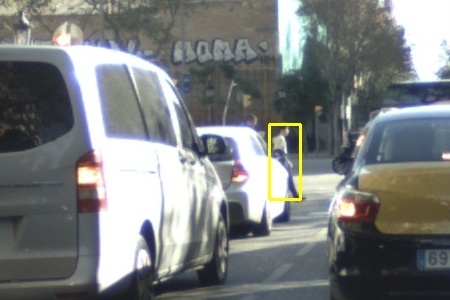}
  \caption{}
  \label{fig:failures:fp:post2}
\end{subfigure}
\begin{subfigure}{0.33\textwidth}
  \centering
  \includegraphics[height=0.15\textheight]{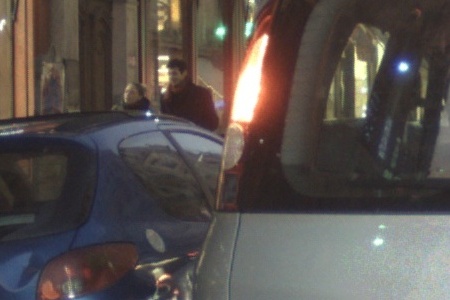}
  \caption{}
  \label{fig:failures:fp:post3}
\end{subfigure}

\begin{subfigure}{0.33\textwidth}
  \centering
  \includegraphics[height=0.15\textheight]{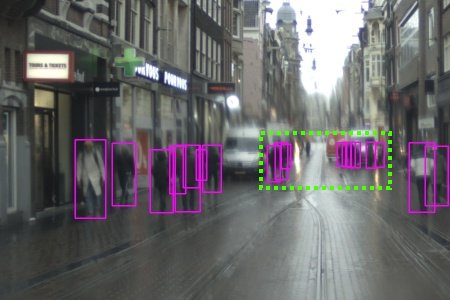}
  \caption{}
  \label{fig:failures:fp:post4}
\end{subfigure}
\begin{subfigure}{0.33\textwidth}
  \centering
  \includegraphics[height=0.15\textheight]{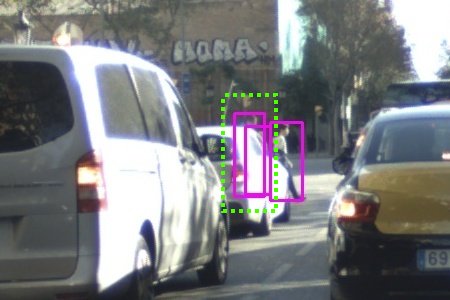}
  \caption{}
  \label{fig:failures:fp:post5}
\end{subfigure}
\begin{subfigure}{0.33\textwidth}
  \centering
  \includegraphics[height=0.15\textheight]{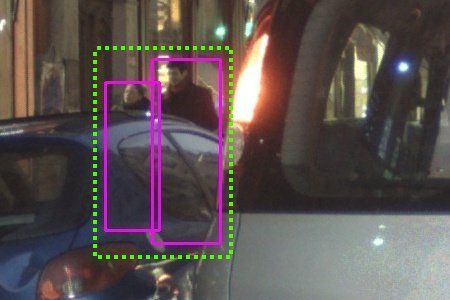}
  \caption{}
  \label{fig:failures:fp:post6}
\end{subfigure}

\begin{subfigure}{0.33\textwidth}
  \centering
  \includegraphics[height=0.15\textheight]{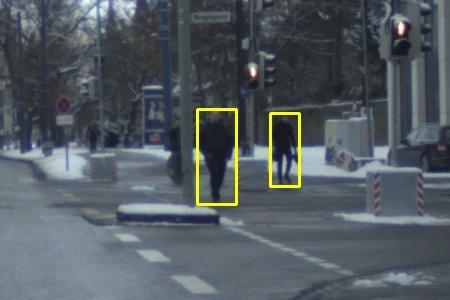}
  \caption{}
  \label{fig:failures:fp:post7}
\end{subfigure}
\begin{subfigure}{0.33\textwidth}
  \centering
  \includegraphics[height=0.15\textheight]{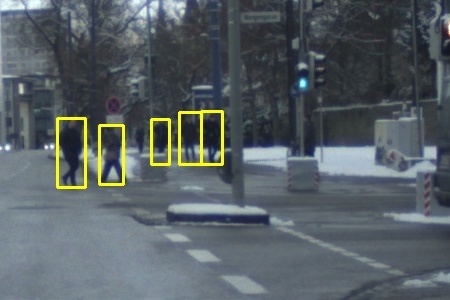}
  \caption{}
  \label{fig:failures:fp:post8}
\end{subfigure}
\begin{subfigure}{0.33\textwidth}
  \centering
  \includegraphics[height=0.15\textheight]{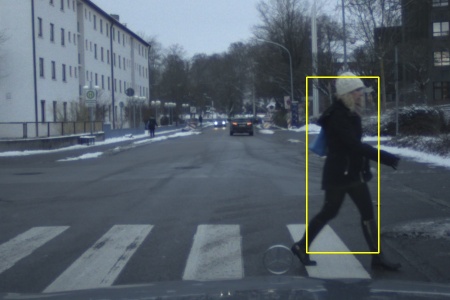}
  \caption{}
  \label{fig:failures:fp:post9}
\end{subfigure}

\begin{subfigure}{0.33\textwidth}
  \centering
  \includegraphics[height=0.15\textheight]{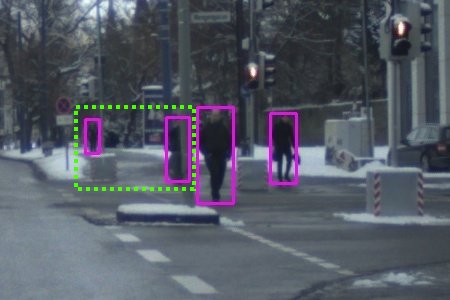}
  \caption{}
  \label{fig:failures:fp:post10}
\end{subfigure}
\begin{subfigure}{0.33\textwidth}
  \centering
  \includegraphics[height=0.15\textheight]{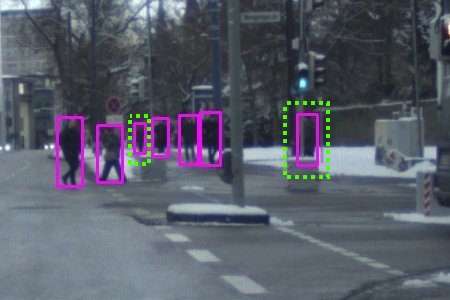}
  \caption{}
  \label{fig:failures:fp:post12}
\end{subfigure}
\begin{subfigure}{0.33\textwidth}
  \centering
  \includegraphics[height=0.15\textheight]{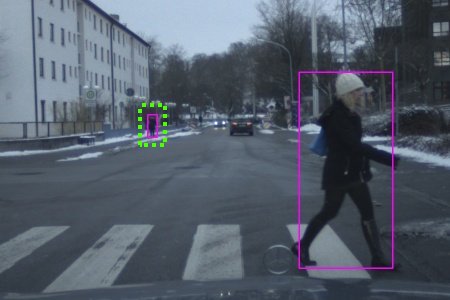}
  \caption{}
  \label{fig:failures:fp:post13}
\end{subfigure}

\caption{Generalization comparison between CSP \cite{Liu2018DBC} with HRNet and Cascade RCNN\cite{cai2019cascade}. Both methods are trained on CityPersons and evaluated on ECP. Images with Yellow bounding box are CSP's detections where as magenta bounding box are Cascade RCNN's detections. Dotted green line illustrates instances where CSP fails to detect.}
\label{fig:gen-com}
\end{figure*}

We show detection quality of \ours{} on different datasets in Figure \ref{fig:good-1}. Top row contain results from \caltecha{} and bottom row show images from \citypersona{} and \ecpa{}. One could observe that \ours{} is robust to crowd density, as presented images show several instances of occlusion, people walking in close vicinity, etc. 
Furthermore, we extracted images from different regions across the globe under varying conditions, shown in Figure, \ref{fig:good-2}. Four different conditions are showcased, for example, Netherlands shows rainy conditions, where people are wearing jackets with hood, rain coats and holding umbrellas. In Italy, summer season is illustrated and people are walking in city-center in close vicinity, often occluding each other. Winter season can be seen in Germany, as snow is visible. Finally, France illustrates a low-illumination scenario, where car headlights can be seen illuminating the scene (also bringing reflections and shadows into play). Aim of this figure is to demonstrate the robustness of \ours{}, since the pre-training is done on web-crawled datasets and thanks to the several diverse and dense scenes, \ours{} has learnt a representation capable of handling several real-world scenarios efficiently. 

Finally, we show qualitatively in Figure \ref{fig:gen-com}, how pedestrian detectors such as CSP lacks in generalization compared with a general object detector such as Cascade RCNN. In the figure \ref{fig:gen-com}, we trained both detectors with HRNet on CityPersons and evaluate on ECP. We picked cases from different cities, under varying lighting conditions (afternoon, evening etc.) and under different weathers such as sunny, raining or snowing. Common failure cases for CSP includes low-illumination/blurry  pedestrians Fig.\ref{fig:gen-com} \textbf{(a)} and \textbf{(d)}, small-scale pedestrians Fig.\ref{fig:gen-com} \textbf{(g)} to \textbf{(l)}, and occlusion Fig.\ref{fig:gen-com} \textbf{(b, c, e, f, k)}. On the contrary, Cascade RCNN seems to be robust to such domain shifts and handles the above mentioned challenging scenarios better than CSP. We argue that probable reasons behind CSP's poor generalizations stems from the fact that it is a single stage detector without feature alignment compared to two-stage detector like Cascade RCNN. Feature alignment improves generalization \cite{chen2019synergistic}. Moreover, two-stage detectors have specific module for feature-alignment (ROI-Align), therefore leading them to more aligned features on unseen domains. This explicit feature alignment is absent in CSP and on unseen domains leads to failure cases, such as occlusion and small-scale pedestrians, where feature alignment becomes pivotal for precise localization. 

\section{How Transformers Fare With Respect To CNN}\label{sec:cnn_v_trans}
In this section, we discuss the results when CNNs are pitched against the recent Transformer Networks. Intuitively, transformer networks do outperform CNN based backbones in direct and cross dataset evaluation. However, when the domain gap is large and we increase the size of the training dataset along with more sources of pre-training, we quantitatively illustrate that CNNs are more robust to domain shifts and have better ability to digest data, compared with transformer networks. To the best of our knowledge, this is one of the first studies that objectively illustrates this.   

In order to make comparison fair, we used the same detector, Cascade R-CNN, in both backbone networks HRNet and Swin Transformer. We included Swin Transformer for evaluation, since, it has achieved state of the art performance on general object detection benchmarks. Therefore, we start by benchmarking in Table \ref{tab:cross-cp-ecp-cal-swin}. It can be observed that Swin Transformer outperforms HRNet on direct and cross-dataset evaluation. Which is intuitive as, Swin Transformer also outperforms HRNet on general object detection as well, thanks to the  \textit{shited window} based self attention which captures a more powerful representation compared with a CNN based backbone such as the HRNet. However, when the domain shift is large, sources of pre-training are fixed to \crowdhumana{} and \widerpersona{} and tested on autonomous driving oriented datasets, we notice the Swin Transformer gets outperformed by HRNet significantly, across all settings as shown in Table\ref{tab:icon-ch-wider-tf}. This finding illustrate that CNNs are more tolerant to domain shifts, especially, if the shifts are not subtle. Inline with the studies in sections above, we test Swin Transformer, through our progressive training pipeline. In table \ref{tab:icon-collapsed-trans}, we pre-train on diverse general person detection dataset and finetune on ECP which is closer to the target domain. Except for the third row, HRNet outperformed Swin Tranformers on all datasets. This trend persists even when we add the target set to the progressive training pipeline as shown in Table \ref{tab:icon-collapsed2-trans}. We attribute this to the fact that potentially one of the main reasons are that the hyper-parameters of transformer networks are not as optmized as they for CNNs. Moreover, transformer networks are also more prone to overfitting compared with the CNNs. Nonetheless, it is still relatively early for transformer networks compared with CNNs, which have been around and used extensively for nearly a decade and their hyperparameters are optimized quite thoroughly.

\begin{table*}[tb]
\centering
\caption{Cross dataset evaluation of HRNet and Swin-Trans. on Autonomous driving benchmarks. Both detectors are trained with Casc. R-CNN as a backbone.}
\label{tab:cross-cp-ecp-cal-swin}
\begin{tabular}{l|c|c|c|c|c}
\hline
Method & Training   & Testing          & Reasonable & Small & Heavy \\ \hline
HRNet & CityPersons & CityPersons  & {11.2}  & 14.0     & {37.0} \\ \hline
Swin-Trans & CityPersons & CityPersons  & \textbf{9.2}  & \textbf{10.5}     & \textbf{36.9} \\ \hline
HRNet  & ECP   & CityPersons&  {10.9} & {11.4}     & 40.9\\ \hline
Swin-Trans & ECP   & CityPersons&  {10.2} & {12.9}     & 39.6\\ \hline
\hline 
HRNet  & ECP    &ECP& {6.9}  & {12.6}     & {33.1} \\ \hline
Swin-Trans & ECP    &ECP& \textbf{4.5}  & \textbf{9.6}     & \textbf{25.2} \\ \hline
HRNet & CityPersons & ECP   &  17.4 & 40.5     & 49.3\\ \hline
Swin-Trans & CityPersons & ECP   &  14.8 & 28.3     & 50.2 \\ \hline
\end{tabular}
\end{table*}


\begin{table*}[!tbh]
\centering
\caption{Cross-dataset evaluation of HRNet and Swin-Trans., when sources of training are fixed to \widerpersona{} and \crowdhumana{}}
\label{tab:icon-ch-wider-tf}
\begin{tabular}{l|c|c|c|c|c}
\hline
Method & Training & Testing             & Reasonable & Small & Heavy \\ \hline
HRNet & \crowdhumana{}& \caltecha{}    & {3.4}  &   {11.2}   & {32.3} \\ \hline
Swin-Trans & \crowdhumana{}& \caltecha{}    &  10.1 & 37.2     & 55.1 \\ \hline
HRNet& \crowdhumana{}& \citypersona{}    & 15.1   & 21.4      & 49.8 \\ \hline
Swin-Trans & \crowdhumana{}& \citypersona{}    & 16.7   &  24.3     & 55.0 \\ \hline
HRNet & \crowdhumana{}& \ecpa{} & 17.9  & 36.5   & 56.9  \\ \hline

Swin-Trans & \crowdhumana{}& \ecpa{} &  21.1 & 42.8  & 56.0   \\ \hline
\hline
HRNet & \widerpersona{}&\caltecha{}    & 3.2  &   10.8   & 31.7 \\ \hline
Swin-Trans & \widerpersona{}&\caltecha{}    & 9.8 &     30.1 & 55.6 \\ \hline
HRNet& \widerpersona{}&\citypersona{}    & 16.0   & 21.6      & 57.4 \\ \hline
Swin-Trans & \widerpersona{}&\citypersona{}    & 13.9   & 32.5 & 57.7 \\ \hline
HRNet & \widerpersona{}&\ecpa{} & 16.1  & 32.8   & 58.0  \\ \hline
Swin-Trans & \widerpersona{}&\ecpa{} &  18.1 & 32.5   & 65.8  \\ \hline
\end{tabular}
\end{table*}


\begin{table*}[tb]
\centering
\caption{Investigating the effect on performance when CrowdHuman, Wider Pedestrian and ECP are merged and both backbones (HRNet and Swin-Trans.) are trained only on the merged dataset.}
\label{tab:icon-collapsed-trans}
\begin{tabular}{l|c|c|c|c|c}
\hline
Method & Training        & Testing     & Reasonable & Small & Heavy \\ \hline
HRNet&CrowdHuman $\rightarrow$ ECP & CP & {10.3}     & {12.6}   & {40.7}  \\ \hline
Swin Trans. &CrowdHuman $\rightarrow$ ECP & CP & {11.0}     & {12.4}   & {43.4}  \\ \hline
HRNet&Wider Pedestrian $\rightarrow$ ECP & CP & {9.7}     & {11.8}  & \textbf{37.7}  \\ \hline
Swin Trans. &Wider Pedestrian $\rightarrow$ ECP & CP & \textbf{9.5}     & \textbf{10.8}  & {39.7}  \\ 
\hline \hline 
HRNet&CrowdHuman $\rightarrow$ ECP & Caltech & {2.9}     & {11.4}   & \textbf{30.8}  \\ \hline
Swin Trans. &CrowdHuman $\rightarrow$ ECP & Caltech & {8.0}     & {28.0}   & {54.4}  \\ \hline
HRNet &Wider Pedestrian $\rightarrow$ ECP & Caltech & \textbf{2.5}     & \textbf{9.9}   & {31.0}  \\ \hline
Swin Trans. &Wider Pedestrian $\rightarrow$ ECP & Caltech & {8.8}     & {28.1}   & {33.9}  \\ \hline
\end{tabular}
\end{table*}

\begin{table*}[tb]
\centering
\caption{Evaluation of HRNet and Swin-Trans. after fine-tuning on the target set.}
\label{tab:icon-collapsed2-trans}
\begin{tabular}{l|c|c|c|c|c}
\hline
Method & Training        & Testing     & Reasonable & Small & Heavy \\ \hline
HRNet&CrowdHuman $\rightarrow$ ECP $\rightarrow$ CP & CP & {8.0}     & {8.5}   & {27.0}  \\ \hline
SwinTrans.&CrowdHuman $\rightarrow$ ECP $\rightarrow$ CP & CP & {9.1}     & {10.0}   & {30.9}  \\ \hline
HRNet&Wider Pedestrian $\rightarrow$ ECP $\rightarrow$ CP & CP & \textbf{7.5}     & \textbf{8.0}  & \textbf{28.0}  \\ \hline
SwinTrans.&Wider Pedestrian $\rightarrow$ ECP $\rightarrow$ CP & CP & {8.9}     & {10.4}  & {33.8}  \\ 
\hline
\end{tabular}
\end{table*}
\section{Finetuning on the Target Set} \label{sec:ft}

Finally, we add the training part of our target set to our progressive training pipeline as illustrated in Table \ref{tab:icon-collapsed2}. Although this paper stresses the importance of cross-dataset evaluation, for the sake of completeness, we also include results for target set fine-tuning. Visible improvement across all splits can be observed for both methods. Interestingly, on CityPersons, CSP gets better than Cascade RCNN. We attribute this to the fact that CSP's design choice is optimized for CityPersons (at the cost of generalization). Therefore, it benefits more than a general design object detector for target set fine-tuning on CityPersons. However, both methods are still comparable and in the case of Caltech, Cascade R-CNN outperforms CSP significantly. Moreover, the performance on Caltech is nearly in the same order, as that of humans (0.88). This also brings to the conclusion that Caltech dataset is almost solved. Next generation pedestrian detectors should use Caltech as a reference, but focus on more challenging dataset such as ECP and CityPersons. Importantly, general object detector such as, Cascade RCNN, without fine-tuning on the target set (cf. Table \ref{tab:icon-collapsed}), can already achieve comparable results to that with fine-tuning as in Table \ref{tab:icon-collapsed2}, making it practical, ready to use and more suitable to real-world problems. 

\begin{table*}[!tbh]
\centering
\caption{After fine-tuning on the target set.}
\label{tab:icon-collapsed2}
\begin{tabular}{l|c|c|c|c|c}
\hline
Method & Training        & Testing     & Reasonable & Small & Heavy \\ \hline
Casc. RCNN&CrowdHuman $\rightarrow$ ECP $\rightarrow$ CP & CP & {8.0}     & {8.5}   & {27.0}  \\ \hline
CSP&CrowdHuman $\rightarrow$ ECP $\rightarrow$ CP & CP & {8.1}     & {8.9}   & {29.3}  \\ \hline
Casc. RCNN&Wider Pedestrian $\rightarrow$ ECP $\rightarrow$ CP & CP & {7.5}     & {8.0}  & {28.0}  \\ \hline
CSP&Wider Pedestrian $\rightarrow$ ECP $\rightarrow$ CP & CP & \textbf{7.1}     & \textbf{7.8}  & \textbf{26.9}  \\ 
\hline \hline 
Casc. RCNN&CrowdHuman $\rightarrow$ ECP $\rightarrow$ Caltech & Caltech & {2.2}     & {8.1}   & {30.7}  \\ \hline
CSP&CrowdHuman $\rightarrow$ ECP $\rightarrow$ Caltech & Caltech & {3.8}     & {9.2}   & {31.2}  \\ \hline
Casc. RCNN &Wider Pedestrian $\rightarrow$ ECP $\rightarrow$ Caltech & Caltech & \textbf{1.7}     & \textbf{7.2}   & \textbf{25.7}  \\ \hline
CSP &Wider Pedestrian $\rightarrow$ ECP $\rightarrow$ Caltech & Caltech & {2.4}     & {7.5}   & {30.3}  \\ \hline
\end{tabular}
\end{table*}

\subsection{Quantitative Results on Leaderboard}

We further evaluated proposed training pipeline along with ensemble of our two models, one pre-trained on \crowdhuman{} and the other one on \widerperson{}. Ensembling is performed by combining the detections followed by non-maxima suppression, using soft-nms \cite{bodla2017soft}. 
Final results are evaluated on the dedicated server\footnote{\url{https://github.com/cvgroup-njust/CityPersons}\\
\url{https://eurocity-dataset.tudelft.nl/eval/benchmarks/detection} \\
\textsuperscript{$\ddagger$}: Correspond to our submissions and use of additional training data\\
\label{ftnote}}
(test set annotations are withheld) of \cityperson{} and \ecp{}, maintained by the benchmark publishers and frequency of submissions are constraint. Moreover, we have included results only for the published methods (detailed evaluations of all methods can be seen on the urls provided in the \emph{footnote} \ref{ftnote}). Results are presented in Table \ref{tab:cp-test} and \ref{tab:ecp-test}. Our submission (Cascade RCNN)  achieves 1st and 2nd on both leaderboards respectively. These results serve as a reference for future methods. However, no other method to the best of our knowledge uses extra training data. Therefore, giving our submissions an unfair advantage. Finally, as stated above, fine-tuning on target set is not the goal of the paper and in many cases it is not practical. In this work, we argue in the favor of cross-dataset evaluation and its importance.

\begin{table}[!tbh]
\centering
\caption{Results on the withheld Test Set of CityPersons.}
\label{tab:cp-test}
 \resizebox{0.98\linewidth}{!}{
\begin{tabular}{l|c|c|c}
\hline
Methods             & Reasonable & Small & Heavy \\ \hline
MS-CNN \cite{cai2016unified}	&13.32 &	15.86 &	51.88 \\ \hline
FRCNN \cite{zhang2017citypersons}	& 12.97 &	37.24	& 50.47 \\ \hline
Cascade MS-CNN \cite{cai2019cascade}	& 11.62 &	13.64 & 47.14 \\ \hline
Repulsion Loss \cite{wang2018repulsion} &	11.48	& 5.67 &	52.59   \\ \hline
AdaptiveNMS \cite{liu2019adaptive} &  11.40     & 13.64  & 46.99   \\ \hline
OR-CNN \cite{zhang2018occlusion}  &  11.32     & 14.19 & 51.43   \\ \hline
HBA-RCNN \cite{lu2019semantic} &  11.26     & 15.68  & 43.7   \\ \hline
MGAN \cite{pang2019maskguided}  &  9.29     & 11.38  & 40.97   \\ \hline
CrowdHuman $\rightarrow$ECP$\rightarrow$CP \textsuperscript{$\ddagger$}   & 8.35     & 9.85  & 27.87   \\ \hline
WiderPerson$\rightarrow$ECP$\rightarrow$CP \textsuperscript{$\ddagger$}   & 8.31     & 10.23  & 28.18   \\ \hline
Ensemble \textsuperscript{$\ddagger$} & {7.69}     & \textbf{9.16}   & \textbf{27.08}  \\ \hline

APD-Pretrain \cite{zhang2020attribute}  & \textbf{7.3}     & {10.8}   & {28.0}  \\ \hline
\end{tabular}
}
\end{table}

\begin{table}[!tbh]
\centering
\caption{Results on the withheld Test Set of ECP.}
\label{tab:ecp-test}
 \resizebox{0.98\linewidth}{!}{
\begin{tabular}{l|c|c|c}
\hline
Training             & Reasonable & Small & Heavy \\ \hline
R-FCN (with OHEM) \cite{braun2018eurocity} &  16.3    & 24.5  & 50.7  \\ \hline
SSD \cite{braun2018eurocity} &  13.1    & 23.5  & 46.0  \\ \hline
Faster R-CNN \cite{braun2018eurocity}  & 10.1     & 19.6  & 38.1   \\ \hline
YOLOv3 \cite{braun2018eurocity} &   9.7   & 18.6  & 40.1  \\ \hline
CrowdHuman$\rightarrow$ECP \textsuperscript{$\ddagger$} & 6.6     & 13.6  & 31.3  \\ \hline
WiderPerson$\rightarrow$ECP \textsuperscript{$\ddagger$}   & {5.5}     & {11.7}   & {26.1}  \\ \hline
APD-Pretrain \cite{zhang2020attribute} & 5.3 & 12 .4 & 26.8 \\\hline
Ensemble \textsuperscript{$\ddagger$}  & {5.1}     & {11.2}   & 25.4  \\ \hline
SPNet \cite{jiang2020sp} & \textbf{4.2}     & \textbf{9.5}   & \textbf{21.6}  \\ \hline
\end{tabular}
}
\end{table}
\section{Conclusions} \label{sec:conc}

In view of the recent developments in the field of pedestrian detection on existing benchmarks for autonomous driving. We thoroughly investigated and evaluated existing state of the art pedestrian detectors, moreover, assessed their practicality and generalization using cross-dataset evaluation as an evaluation protocol. Through this study, we have pointed out an important, but often overlooked conclusion, that is, the current pedestrian detectors lack generalisation and do not handle well even small domain shifts.
We have attributed lack of generalization to the fact that current pedestrian detectors are tailored towards target datasets, and their architectural design contains strong biasness towards the target domain, at the cost of generalisation. Therefore, we have illustrated that general object detectors, are more robust and are better at handling domain shifts due to generic design. Besides detectors, we also investigated the role of backbone-network architectures in generalization. We have compared the generalization ability of convolutional neural networks and the recent transformer networks, using the same detection architecture (Cascade R-CNN). We have shown that, although transformers out perform CNNs in direct dataset evaluation, still, the generalization ability of CNNs is better than that of transformer networks. Moreover, CNNs have shown to be better at digesting largescale datasets than transformers, potentially due to the fact that CNN's hyper-parameters are thoroughly optimized over decade of rigorous research, and it is still relatively early days for transformer networks. Additionally, we have proposed a progressive-training pipeline in this paper, where we have shown that benchmarks diverse in scenes and dense in pedestrians, serves to be an efficient source of pre-training, especially in the context of autonomous driving, where most benchmarks are monotonous in nature. Finally, findings presented in this paper can inspire the next generation of pedestrian detectors, which are more generic, practical and are ready to use.

\bibliographystyle{ieee}
\bibliography{main}

\begin{thebibliography}{10}\itemsep=-1pt

\bibitem{angelova2015real}
A.~Angelova, A.~Krizhevsky, V.~Vanhoucke, A.~Ogale, and D.~Ferguson.
\newblock Real-time pedestrian detection with deep network cascades.
\newblock 2015.

\bibitem{benfold2011stable}
B.~Benfold and I.~Reid.
\newblock Stable multi-target tracking in real-time surveillance video.
\newblock In {\em CVPR 2011}, pages 3457--3464. IEEE, 2011.

\bibitem{bodla2017soft}
N.~Bodla, B.~Singh, R.~Chellappa, and L.~S. Davis.
\newblock Soft-nms--improving object detection with one line of code.
\newblock In {\em Proceedings of the IEEE international conference on computer
  vision}, pages 5561--5569, 2017.

\bibitem{braun2018eurocity}
M.~Braun, S.~Krebs, F.~Flohr, and D.~M. Gavrila.
\newblock Eurocity persons: A novel benchmark for person detection in traffic
  scenes.
\newblock {\em IEEE transactions on pattern analysis and machine intelligence},
  41(8):1844--1861, 2019.

\bibitem{brazil2017illuminating}
G.~Brazil, X.~Yin, and X.~Liu.
\newblock Illuminating pedestrians via simultaneous detection \& segmentation.
\newblock In {\em Proceedings of the IEEE International Conference on Computer
  Vision}, pages 4950--4959, 2017.

\bibitem{cai2016unified}
Z.~Cai, Q.~Fan, R.~S. Feris, and N.~Vasconcelos.
\newblock A unified multi-scale deep convolutional neural network for fast
  object detection.
\newblock In {\em European conference on computer vision}, pages 354--370.
  Springer, 2016.

\bibitem{cai2015learning}
Z.~Cai, M.~Saberian, and N.~Vasconcelos.
\newblock Learning complexity-aware cascades for deep pedestrian detection.
\newblock In {\em Proceedings of the IEEE International Conference on Computer
  Vision}, pages 3361--3369, 2015.

\bibitem{cai2019cascade}
Z.~Cai and N.~Vasconcelos.
\newblock Cascade r-cnn: High quality object detection and instance
  segmentation.
\newblock {\em IEEE Transactions on Pattern Analysis and Machine Intelligence},
  2019.

\bibitem{campmany2016gpu}
V.~Campmany, S.~Silva, A.~Espinosa, J.~C. Moure, D.~V{\'a}zquez, and A.~M.
  L{\'o}pez.
\newblock Gpu-based pedestrian detection for autonomous driving.
\newblock {\em Procedia Computer Science}, 80:2377--2381, 2016.

\bibitem{chen2019synergistic}
C.~Chen, Q.~Dou, H.~Chen, J.~Qin, and P.-A. Heng.
\newblock Synergistic image and feature adaptation: Towards cross-modality
  domain adaptation for medical image segmentation.
\newblock In {\em Proceedings of the AAAI Conference on Artificial
  Intelligence}, volume~33, pages 865--872, 2019.

\bibitem{dalal2005histograms}
N.~Dalal and B.~Triggs.
\newblock Histograms of oriented gradients for human detection.
\newblock In {\em international Conference on computer vision \& Pattern
  Recognition (CVPR'05)}, volume~1, pages 886--893. IEEE Computer Society,
  2005.

\bibitem{dollar2014fast}
P.~Doll{\'a}r, R.~Appel, S.~Belongie, and P.~Perona.
\newblock Fast feature pyramids for object detection.
\newblock {\em IEEE Transactions on Pattern Analysis and Machine Intelligence},
  36(8):1532--1545, 2014.

\bibitem{dollar2012pedestrian}
P.~Dollar, C.~Wojek, B.~Schiele, and P.~Perona.
\newblock Pedestrian detection: An evaluation of the state of the art.
\newblock {\em IEEE transactions on pattern analysis and machine intelligence},
  34(4):743--761, 2012.

\bibitem{ess2007depth}
A.~Ess, B.~Leibe, and L.~Van~Gool.
\newblock Depth and appearance for mobile scene analysis.
\newblock In {\em 2007 IEEE 11th international conference on computer vision},
  pages 1--8. IEEE, 2007.

\bibitem{geiger2012we}
A.~Geiger, P.~Lenz, and R.~Urtasun.
\newblock Are we ready for autonomous driving? the kitti vision benchmark
  suite.
\newblock In {\em 2012 IEEE Conference on Computer Vision and Pattern
  Recognition}, pages 3354--3361. IEEE, 2012.

\bibitem{girshick2014rich}
R.~Girshick, J.~Donahue, T.~Darrell, and J.~Malik.
\newblock Rich feature hierarchies for accurate object detection and semantic
  segmentation.
\newblock In {\em Proceedings of the IEEE conference on computer vision and
  pattern recognition}, pages 580--587, 2014.

\bibitem{hattori2015learning}
H.~Hattori, V.~Naresh~Boddeti, K.~M. Kitani, and T.~Kanade.
\newblock Learning scene-specific pedestrian detectors without real data.
\newblock In {\em Proceedings of the IEEE Conference on Computer Vision and
  Pattern Recognition}, pages 3819--3827, 2015.

\bibitem{hbaieb2019pedestrian}
A.~Hbaieb, J.~Rezgui, and L.~Chaari.
\newblock Pedestrian detection for autonomous driving within cooperative
  communication system.
\newblock In {\em 2019 IEEE Wireless Communications and Networking Conference
  (WCNC)}, pages 1--6. IEEE, 2019.

\bibitem{he2017mask}
K.~He, G.~Gkioxari, P.~Doll{\'a}r, and R.~Girshick.
\newblock Mask r-cnn.
\newblock In {\em Proceedings of the IEEE international conference on computer
  vision}, pages 2961--2969, 2017.

\bibitem{hosang2015taking}
J.~Hosang, M.~Omran, R.~Benenson, and B.~Schiele.
\newblock Taking a deeper look at pedestrians.
\newblock In {\em Proceedings of the IEEE Conference on Computer Vision and
  Pattern Recognition}, pages 4073--4082, 2015.

\bibitem{howard2017mobilenets}
A.~G. Howard, M.~Zhu, B.~Chen, D.~Kalenichenko, W.~Wang, T.~Weyand,
  M.~Andreetto, and H.~Adam.
\newblock Mobilenets: Efficient convolutional neural networks for mobile vision
  applications.
\newblock {\em arXiv preprint arXiv:1704.04861}, 2017.

\bibitem{jiang2020sp}
C.~Jiang, H.~Xu, W.~Zhang, X.~Liang, and Z.~Li.
\newblock Sp-nas: Serial-to-parallel backbone search for object detection.
\newblock In {\em Proceedings of the IEEE/CVF Conference on Computer Vision and
  Pattern Recognition}, pages 11863--11872, 2020.

\bibitem{li2020box}
J.~Li, S.~Liao, H.~Jiang, and L.~Shao.
\newblock Box guided convolution for pedestrian detection.
\newblock In {\em Proceedings of the 28th ACM International Conference on
  Multimedia}, pages 1615--1624, 2020.

\bibitem{lin2017feature}
T.-Y. Lin, P.~Doll{\'a}r, R.~Girshick, K.~He, B.~Hariharan, and S.~Belongie.
\newblock Feature pyramid networks for object detection.
\newblock In {\em Proceedings of the IEEE conference on computer vision and
  pattern recognition}, pages 2117--2125, 2017.

\bibitem{liu2019adaptive}
S.~Liu, D.~Huang, and Y.~Wang.
\newblock Adaptive nms: Refining pedestrian detection in a crowd.
\newblock In {\em Proceedings of the IEEE Conference on Computer Vision and
  Pattern Recognition}, pages 6459--6468, 2019.

\bibitem{liu2016ssd}
W.~Liu, D.~Anguelov, D.~Erhan, C.~Szegedy, S.~Reed, C.-Y. Fu, and A.~C. Berg.
\newblock Ssd: Single shot multibox detector.
\newblock In {\em European conference on computer vision}, pages 21--37.
  Springer, 2016.

\bibitem{liu2019center}
W.~Liu, I.~Hasan, and S.~Liao.
\newblock Center and scale prediction: A box-free approach for pedestrian and
  face detection.
\newblock {\em arXiv preprint arXiv:1904.02948}, 2019.

\bibitem{liu2018learning}
W.~Liu, S.~Liao, W.~Hu, X.~Liang, and X.~Chen.
\newblock Learning efficient single-stage pedestrian detectors by asymptotic
  localization fitting.
\newblock In {\em Proceedings of the European Conference on Computer Vision
  (ECCV)}, pages 618--634, 2018.

\bibitem{Liu2018DBC}
W.~Liu, S.~Liao, W.~Ren, W.~Hu, and Y.~Yu.
\newblock High-level semantic feature detection: A new perspective for
  pedestrian detection.
\newblock In {\em Proceedings of the IEEE Conference on Computer Vision and
  Pattern Recognition}, 2019.

\bibitem{liu2021swin}
Z.~Liu, Y.~Lin, Y.~Cao, H.~Hu, Y.~Wei, Z.~Zhang, S.~Lin, and B.~Guo.
\newblock Swin transformer: Hierarchical vision transformer using shifted
  windows.
\newblock {\em arXiv preprint arXiv:2103.14030}, 2021.

\bibitem{zhang2019widerperson}
C.~C. Loy, D.~Lin, W.~Ouyang, Y.~Xiong, S.~Yang, Q.~Huang, D.~Zhou, W.~Xia,
  Q.~Li, P.~Luo, et~al.
\newblock Wider face and pedestrian challenge 2018: Methods and results.
\newblock {\em arXiv preprint arXiv:1902.06854}, 2019.

\bibitem{lu2019semantic}
R.~Lu and H.~Ma.
\newblock Semantic head enhanced pedestrian detection in a crowd.
\newblock {\em arXiv preprint arXiv:1911.11985}, 2019.

\bibitem{mahajan2018exploring}
D.~Mahajan, R.~Girshick, V.~Ramanathan, K.~He, M.~Paluri, Y.~Li, A.~Bharambe,
  and L.~van~der Maaten.
\newblock Exploring the limits of weakly supervised pretraining.
\newblock In {\em Proceedings of the European Conference on Computer Vision
  (ECCV)}, pages 181--196, 2018.

\bibitem{mao2017can}
J.~Mao, T.~Xiao, Y.~Jiang, and Z.~Cao.
\newblock What can help pedestrian detection?
\newblock In {\em Proceedings of the IEEE Conference on Computer Vision and
  Pattern Recognition}, pages 3127--3136, 2017.

\bibitem{munder2006experimental}
S.~Munder and D.~M. Gavrila.
\newblock An experimental study on pedestrian classification.
\newblock {\em IEEE transactions on pattern analysis and machine intelligence},
  28(11):1863--1868, 2006.

\bibitem{paisitkriangkrai2014strengthening}
S.~Paisitkriangkrai, C.~Shen, and A.~Van Den~Hengel.
\newblock Strengthening the effectiveness of pedestrian detection with
  spatially pooled features.
\newblock In {\em European Conference on Computer Vision}, pages 546--561.
  Springer, 2014.

\bibitem{pang2019maskguided}
Y.~Pang, J.~Xie, M.~H. Khan, R.~M. Anwer, F.~S. Khan, and L.~Shao.
\newblock Mask-guided attention network for occluded pedestrian detection,
  2019.

\bibitem{ren2015faster}
S.~Ren, K.~He, R.~Girshick, and J.~Sun.
\newblock Faster r-cnn: Towards real-time object detection with region proposal
  networks.
\newblock In {\em Advances in neural information processing systems}, pages
  91--99, 2015.

\bibitem{shao2018crowdhuman}
S.~Shao, Z.~Zhao, B.~Li, T.~Xiao, G.~Yu, X.~Zhang, and J.~Sun.
\newblock Crowdhuman: A benchmark for detecting human in a crowd.
\newblock {\em arXiv preprint arXiv:1805.00123}, 2018.

\bibitem{simonyan2014very}
K.~Simonyan and A.~Zisserman.
\newblock Very deep convolutional networks for large-scale image recognition.
\newblock {\em arXiv preprint arXiv:1409.1556}, 2014.

\bibitem{song2020progressive}
X.~Song, K.~Zhao, W.-S. C.~H. Zhang, and J.~Guo.
\newblock Progressive refinement network for occluded pedestrian detection.
\newblock In {\em Proc. European Conference on Computer Vision}, volume~7,
  page~9, 2020.

\bibitem{sun2018fishnet}
S.~Sun, J.~Pang, J.~Shi, S.~Yi, and W.~Ouyang.
\newblock Fishnet: A versatile backbone for image, region, and pixel level
  prediction.
\newblock In {\em Advances in Neural Information Processing Systems}, pages
  760--770, 2018.

\bibitem{viola2004robust}
P.~Viola and M.~J. Jones.
\newblock Robust real-time face detection.
\newblock {\em International journal of computer vision}, 57(2):137--154, 2004.

\bibitem{wang2019deep}
J.~Wang, K.~Sun, T.~Cheng, B.~Jiang, C.~Deng, Y.~Zhao, D.~Liu, Y.~Mu, M.~Tan,
  X.~Wang, et~al.
\newblock Deep high-resolution representation learning for visual recognition.
\newblock {\em arXiv preprint arXiv:1908.07919}, 2019.

\bibitem{wang2018repulsion}
X.~Wang, T.~Xiao, Y.~Jiang, S.~Shao, J.~Sun, and C.~Shen.
\newblock Repulsion loss: detecting pedestrians in a crowd.
\newblock In {\em Proceedings of the IEEE Conference on Computer Vision and
  Pattern Recognition}, pages 7774--7783, 2018.

\bibitem{wojek2009multi}
C.~Wojek, S.~Walk, and B.~Schiele.
\newblock Multi-cue onboard pedestrian detection.
\newblock In {\em 2009 IEEE Conference on Computer Vision and Pattern
  Recognition}, pages 794--801. IEEE, 2009.

\bibitem{wu2007cluster}
B.~Wu and R.~Nevatia.
\newblock Cluster boosted tree classifier for multi-view, multi-pose object
  detection.
\newblock In {\em 2007 IEEE 11th International Conference on Computer Vision},
  pages 1--8. IEEE, 2007.

\bibitem{xie2017aggregated}
S.~Xie, R.~Girshick, P.~Doll{\'a}r, Z.~Tu, and K.~He.
\newblock Aggregated residual transformations for deep neural networks.
\newblock In {\em Proceedings of the IEEE conference on computer vision and
  pattern recognition}, pages 1492--1500, 2017.

\bibitem{zhang2020attribute}
J.~Zhang, L.~Lin, J.~Zhu, Y.~Li, Y.-c. Chen, Y.~Hu, and C.~S. Hoi.
\newblock Attribute-aware pedestrian detection in a crowd.
\newblock {\em IEEE Transactions on Multimedia}, 2020.

\bibitem{zhang2016faster}
L.~Zhang, L.~Lin, X.~Liang, and K.~He.
\newblock Is faster r-cnn doing well for pedestrian detection?
\newblock In {\em European conference on computer vision}, pages 443--457.
  Springer, 2016.

\bibitem{zhang2016far}
S.~Zhang, R.~Benenson, M.~Omran, J.~Hosang, and B.~Schiele.
\newblock How far are we from solving pedestrian detection?
\newblock In {\em Proceedings of the IEEE Conference on Computer Vision and
  Pattern Recognition}, pages 1259--1267, 2016.

\bibitem{zhang2017citypersons}
S.~Zhang, R.~Benenson, and B.~Schiele.
\newblock Citypersons: A diverse dataset for pedestrian detection.
\newblock In {\em Proceedings of the IEEE Conference on Computer Vision and
  Pattern Recognition}, pages 3213--3221, 2017.

\bibitem{zhang2018occlusion}
S.~Zhang, L.~Wen, X.~Bian, Z.~Lei, and S.~Z. Li.
\newblock Occlusion-aware r-cnn: detecting pedestrians in a crowd.
\newblock In {\em Proceedings of the European Conference on Computer Vision
  (ECCV)}, pages 637--653, 2018.

\bibitem{zhang2019widerpersondataset}
S.~Zhang, Y.~Xie, J.~Wan, H.~Xia, S.~Z. Li, and G.~Guo.
\newblock Widerperson: A diverse dataset for dense pedestrian detection in the
  wild.
\newblock {\em IEEE Transactions on Multimedia}, 2019.

\bibitem{zhou2018bi}
C.~Zhou and J.~Yuan.
\newblock Bi-box regression for pedestrian detection and occlusion estimation.
\newblock In {\em Proceedings of the European Conference on Computer Vision
  (ECCV)}, pages 135--151, 2018.

\end{thebibliography}

\end{document}